\pdfoutput=1
\documentclass[10pt]{article} %
\usepackage[accepted]{tmlr}

\usepackage[utf8]{inputenc} %
\usepackage[T1]{fontenc}    %
\usepackage{hyperref}       %
\usepackage{url}            %
\usepackage{booktabs}       %
\usepackage{amsfonts}       %
\usepackage{nicefrac}       %
\usepackage{microtype}      %
\usepackage{xcolor}         %

\let\tmlrAND\AND
\let\AND\relax

\usepackage{amsmath}
\usepackage{amssymb}
\usepackage{mathtools}
\usepackage{amsthm}
\theoremstyle{plain}
\newtheorem{theorem}{Theorem}[section]
\newtheorem{proposition}[theorem]{Proposition}

\theoremstyle{definition}

\theoremstyle{remark}

\usepackage{bm}
\usepackage{graphicx}
\usepackage{subcaption}
\usepackage{float}
\usepackage{wrapfig}
\usepackage{array}
\usepackage{multirow}
\usepackage{dsfont}
\usepackage{cancel}
\usepackage{soul}
\usepackage[normalem]{ulem}
\usepackage{siunitx}
\usepackage{algorithm}
\usepackage{algorithmic}
\usepackage[page,toc,titletoc,title]{appendix}
\usepackage[capitalize,noabbrev]{cleveref}
\usepackage{colortbl}
\usepackage{tabularx}
\usepackage{tcolorbox}
\usepackage{mdframed}
\usepackage{pifont}
\usepackage{xargs}
\usepackage{cases}

\hypersetup{
	final,
	colorlinks,
	linkcolor=ourred,
	citecolor=ourblue,
	urlcolor=url,
}

\definecolor{ourblue}{rgb}{0.368,0.507,0.71}
\definecolor{ourorange}{rgb}{0.881,0.611,0.142}
\definecolor{ourgreen}{rgb}{0.56,0.692,0.195}
\definecolor{ourred}{rgb}{0.923,0.386,0.209}
\definecolor{ourviolet}{rgb}{0.528,0.471,0.701}
\definecolor{ourbrown}{rgb}{0.772,0.432,0.102}
\definecolor{ourlightblue}{rgb}{0.364,0.619,0.782}
\definecolor{ourdarkgreen}{rgb}{0.572,0.586,0.}

\definecolor{ourcyan2}{rgb}{0.125,0.722,0.804}
\definecolor{ourred2}{rgb}{0.863,0.184,0.047}
\definecolor{ouryellow2}{cmyk}{0,0.16,1.0,0.07}
\definecolor{ourviolet2}{cmyk}{0.55,0.56,0,0.47}
\definecolor{ourorange2}{cmyk}{0,0.46,0.89,0.11}

\definecolor{discretecolor}{RGB}{11,83,150}
\definecolor{gaussiancolor}{RGB}{230,145,56}
\definecolor{argmaxcolor}{RGB}{154,0,0}
\definecolor{grayseq}{RGB}{120,120,120}
\definecolor{url}{HTML}{c55a3a}
\definecolor{ourslightgray}{RGB}{235, 235, 235}

\newcommand{\Fig}[1]{Figure~\ref{#1}}

\newcommand{\tab}[1]{Table~\ref{#1}}
\newcommand{\Eqn}[1]{(\ref{#1})}

\renewcommand{\sec}[1]{Sec.~\ref{#1}}
\newcommand{\supp}[1]{Suppl.~\ref{#1}}

\newcommand{\algo}[1]{Algo.~\ref{#1}}

\newcommand{\method}{$\mathbb{S}$-FLM\xspace}
\newcommand{\arch}{$\mathbb{S}$-arch\xspace}

\newcommand{\mask}{[\textsc{mask}]~}
\newcommand{\bostoken}{[\textsc{BOS}]~}
\newcommand{\septoken}{[\textsc{SEP}]~}

\usepackage{xspace}
\makeatletter
\DeclareRobustCommand\onedot{\futurelet\@let@token\@onedot}
\def\@onedot{\ifx\@let@token.\else.\null\fi\xspace}

\makeatother

\newcommand{\Comment}[1]{\hfill{\color{gray}\footnotesize$\triangleright$~#1}}

\def\x{{\mathbf x}}

\def\supl{{\color{grayseq}\ell}}

\def\month{09}
\def\year{2026}
\def\openreview{\url{https://openreview.net/forum?id=jBh2WNssnL}}

\title{Language Modeling with Hyperspherical Flows}

\author{\name Justin Deschenaux \email justin.deschenaux@gmail.com \\
      \addr EPFL
      \tmlrAND
      \name Caglar Gulcehre \\
      \addr EPFL, Microsoft AI}

\begin{document}

\maketitle

\begin{abstract}
Discrete Diffusion Language Models have progressed rapidly as an alternative to Autoregressive (AR) models, motivated by their ability to generate tokens in parallel. However, for tractability, discrete diffusion models sample from a factorized distribution, which is less expressive than the AR factorization. Recent Flow Language Models (FLMs) apply continuous flows to language, transporting noise to data with a deterministic ODE that avoids factorized sampling.
FLMs operate on one-hot vectors whose dimension scales with the vocabulary size, making FLMs costly to train. Moreover, since all distinct one-hot vectors are equidistant in $\ell_2$, adding Gaussian noise has no clear semantic interpretation (unlike images, where Gaussian noise progressively degrades structure).
We introduce \method, a latent FLM on the hypersphere. \method generates sequences by rotating vectors on $\mathbb{S}^{d-1}$ along a velocity field learned with Cross-Entropy, avoiding the overhead of materializing one-hot vectors.
Previous FLMs match AR in Generative Perplexity (Gen.\ PPL), but samples with high likelihood are not necessarily correct in verifiable domains such as math and code. \method substantially improves over prior FLMs on large-vocabulary reasoning, increasing the GSM8K accuracy from less than 1\% to 12--18\%, depending on the decoding strategy. \method matches discrete diffusion models (MDLM and Duo) under standard-temperature sampling ($T=1$), but a gap remains under optimized low-temperature ($T=0.1$) decoding. Our code is available at \url{https://github.com/jdeschena/s-flm}.
\end{abstract}
\begin{figure}[H]
    \centering
    \begin{subfigure}[t]{0.49\textwidth}
        \centering
        \includegraphics[width=\linewidth]{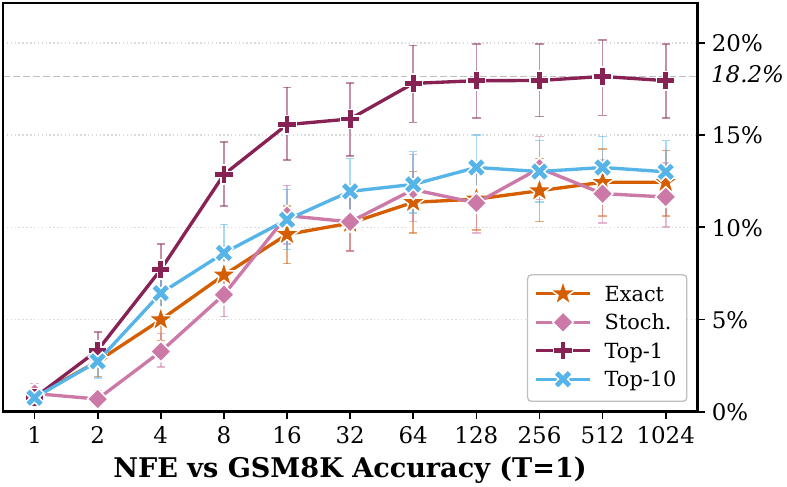}
    \end{subfigure}\hfill
    \begin{subfigure}[t]{0.49\textwidth}
        \centering
        \includegraphics[width=\linewidth]{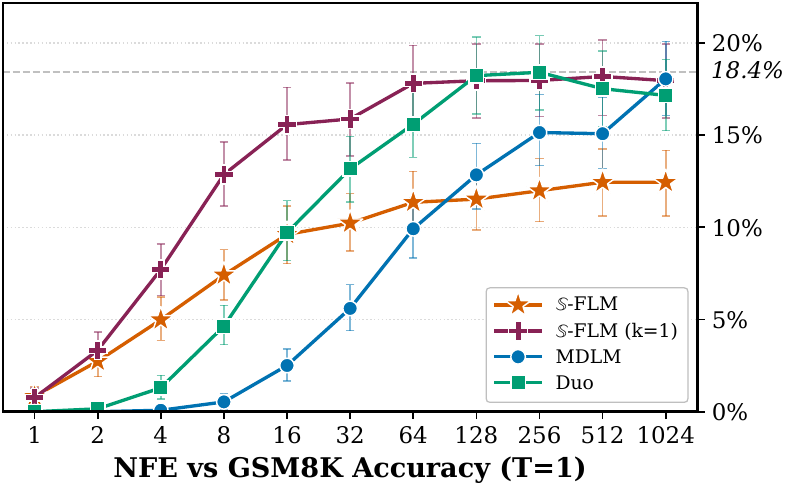}
    \end{subfigure}
    \caption{\textbf{Accuracy on GSM8K} at $T = 1$. \textbf{Left:} Decoding strategies for \method with the \arch (\sec{sec:hyperspherical-architecture}). Exact-velocity \Eqn{eq:flm-marginal-vf} and stochastic decoding (\algo{alg:sampling-stochastic}, \emph{Stoch.}) plateau near $12\%$. Restricting the velocity to the top-$k$ entries of $p^\theta_{1|t}$ improves the accuracy, with top-$1$ reaching $\sim18\%$. \textbf{Right:} \method (with the \arch) vs.\ MDLM and Duo. With the exact velocity, \method beats both baselines at NFE $\leq 16$.}
    \label{fig:tinygsm-T1p0}
\end{figure}
\begin{figure}[t]
    \centering
    \includegraphics[width=\linewidth]{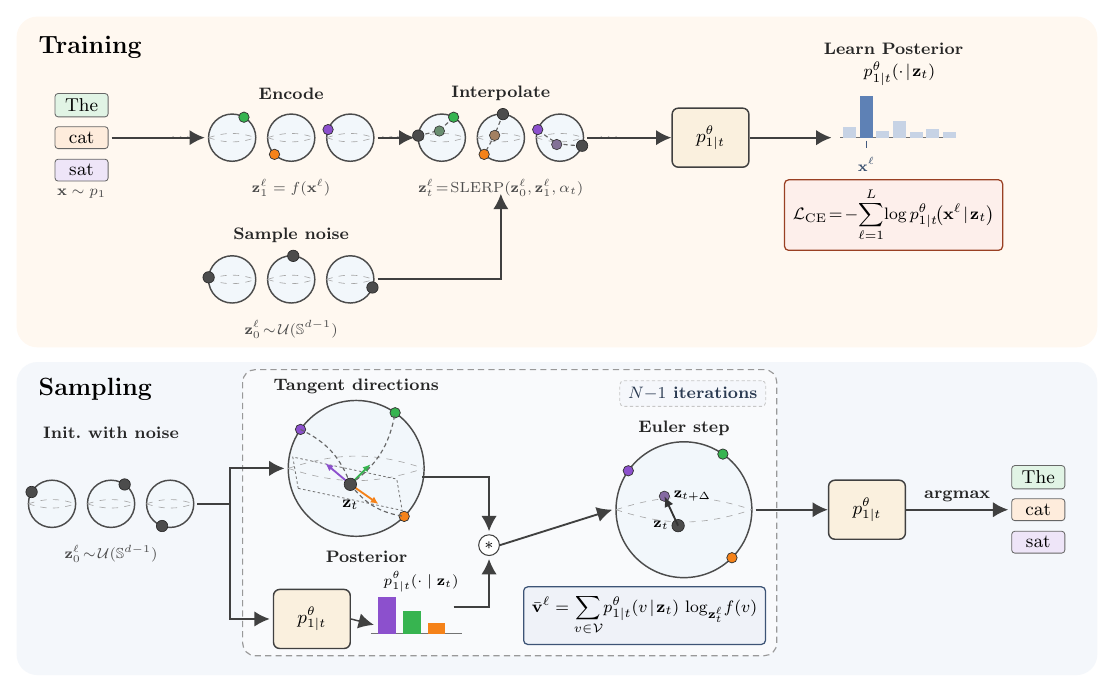}
    \caption{\textbf{\method overview.} \textbf{Training (top):} we embed each token as a unit-norm vector on $\mathbb{S}^{d-1}$. We obtain the noisy latent $\mathbf{z}_t^\supl$ by SLERP between the clean embedding and a random vector on $\mathbb{S}^{d-1}$. We train the denoiser $p^\theta_{1|t}$ with Cross-Entropy. \textbf{Sampling (bottom):} $p^\theta_{1|t}$ defines a velocity field by marginalizing over tangent vectors pointing toward each clean embedding $\hat{\mathbf{e}}_v$, $v \in \mathcal{V}$. Starting from uniform noise on $\mathbb{S}^{d-1}$, we integrate along the velocity field and decode the final latent via $\arg\max_{v \in \mathcal{V}} p^\theta_{1|1}(v \mid \mathbf{z}_1)$.}
    \label{fig:S-FLM-fig1}
\end{figure}
\section{Introduction}
Autoregressive (AR) models currently dominate language modeling. Thanks to the chain-rule factorization and the Transformer architecture \citep{vaswani2017attentionneed}, the AR likelihood is fast to evaluate, and AR language models scale to large sizes \citep{kaplan2020scalinglawsneurallanguage, openai2024gpt4technicalreport, openai2024gptoss, grattafiori2024llama3herdmodels, geminiteam2025gemini, gemmateam2025gemma3technicalreport}. However, during sampling, AR models need one forward pass per token, and causal attention can hurt performance on reasoning tasks that require bidirectional context \citep{papadopoulos2024arrowstimelargelanguage, kitouni2024factorizationcursetokenspredict, zhangli2024reversenumberdecodingorder, nagarajan2025roll}.

Discrete diffusion models \citep{austin2023structureddenoisingdiffusionmodels, campbell2022continuoustimeframeworkdiscrete, sahoo2024simpleeffectivemaskeddiffusion, gat2024discreteflowmatching, sahoo2025diffusionduality, shi2025simplifiedgeneralizedmaskeddiffusion, nie2025scalingmaskeddiffusionmodels, vonruette2026scalingbehaviordiscretediffusion, sahoo2026scalingmaskeddiffusionlanguage, wu2025fastdllmtrainingfreeaccelerationdiffusion} approach AR models in Generative Perplexity (Gen.\ PPL) while generating in parallel and using bidirectional context. However, at each denoising step, these models \emph{sample} tokens from factorized marginals rather than jointly. The factorization makes discrete diffusion less expressive than AR models when generating tokens in parallel.

Continuous flows trained with Flow Matching \citep{lipman2023flowmatchinggenerativemodeling, liu2022flowstraightfastlearning, albergo2023buildingnormalizingflowsstochastic} learn a velocity field that defines an \emph{Ordinary Differential Equation} (ODE) whose solution transports noisy samples to the data distribution. Thus, inference steps update all positions jointly and avoid the factorized sampling issue of discrete diffusion.
Recent work \citep{roos2026categoricalflowmaps, lee2026onesteplanguagemodelingcontinuous, potaptchik2026discreteflowmaps} revived interest in flow-based language models \citep{li2022diffusionlmimprovescontrollabletext, dieleman2022continuousdiffusioncategoricaldata, gulrajani2023likelihoodbaseddiffusionlanguagemodels}, now called \emph{Flow Language Models} (FLMs). Recent FLMs represent tokens as one-hot vectors, add Gaussian noise, and train a denoiser with Cross-Entropy (CE). Although they match the Gen.\ PPL of AR and discrete diffusion models, these FLMs have two main shortcomings. \textbf{(1)} Representing tokens as one-hot vectors is costly. \emph{Large Language Models} (LLMs) commonly use vocabularies containing 100k--200k tokens \citep{openai2024gptoss, qwen2025qwen25technicalreport}, so large FLMs would need to store a $>100$k-dimensional vector for every token. Once Gaussian noise is added to these one-hot vectors, the denoiser must multiply them with the embedding matrix instead of looking up a single
vector. Therefore, FLMs are slower to train than discrete diffusion and AR models. \textbf{(2)} In images, Gaussian diffusion smoothly degrades the higher-frequency components first. For one-hot vectors, adding Gaussian noise has no such clear interpretation.
\paragraph{Contributions.}
We propose the \emph{Hyperspherical Flow Language Model} (\method), a flow over embeddings that does not need to materialize one-hot vectors. \textbf{(1)} Recall that the cosine distance captures the similarity between token embeddings better than the Euclidean distance \citep{mikolov2013efficientestimationwordrepresentations, pennington2014glove, wang2020understandingcontrastiverepresentationlearning}. Since the cosine distance is determined by the arc length on the unit hypersphere, we implement \method as a Riemannian flow on $\mathbb{S}^{d-1}$. Our forward process transports unit-norm embeddings toward a uniform prior. \textbf{(2)} \method operates on $d$-dimensional embeddings rather than $|\mathcal{V}|$-dimensional one-hot vectors. Thus, assuming the same backbone, \method has a training cost similar to that of discrete diffusion models (unlike FLMs on one-hot vectors, which are more costly). We further introduce \arch, a backbone whose activations lie on $\mathbb{S}^{d-1}$. Keeping the activations on the same hypersphere as the inputs improves the sample quality on GSM8K and OpenWebText (OWT) \citep{Gokaslan2019OpenWeb}. \textbf{(3)} On GSM8K, the accuracy of previous FLMs trained on TinyGSM \citep{liu2023tinygsm} is less than $1\%$. In contrast, \method reaches $\sim 12\%$ with the exact velocity and $\sim 18\%$ with the top-$1$ velocity (\sec{par:top-k-velocity}). At standard temperature ($T = 1$), \method with the top-$1$ velocity closes the gap to MDLM and Duo across \emph{Number of Function Evaluations} (NFE) budgets, and \method with the exact velocity outperforms them at NFE $\leq 16$ (\Fig{fig:tinygsm-T1p0}). A gap remains at low-temperature ($T = 0.1$) decoding, where MDLM and Duo reach $33$--$36\%$ (\Fig{fig:tinygsm-T0p1}).
\section{Background}
\paragraph{Notation.}
We denote by $\mathcal{V}$ the finite vocabulary of size $|\mathcal{V}|$. Boldface letters denote either sequences $\x \in \mathcal{V}^L$ of $L$ tokens, with $\x^\supl$ the $\ell$-th element, or vectors in $\mathbb{R}^d$. The meaning is clear from context. We write $\mathbb{S}^{d-1} := \{\x \in \mathbb{R}^d : \|\x\| = 1\}$ for the unit hypersphere in $\mathbb{R}^d$, and $\mathcal{U}(\mathbb{S}^{d-1})$ for the uniform distribution on $\mathbb{S}^{d-1}$. The token embeddings are stored in a lookup table $\mathbf{E} \in \mathbb{R}^{|\mathcal{V}| \times d}$, and we write $\mathbf{e}_v \in \mathbb{R}^d$ for the row associated with $v \in \mathcal{V}$. We write $\zeta_v$ for the logit that the denoiser $p^\theta_{1|t}$ (\sec{sec:method}) assigns to token $v$.
\paragraph{Language Modeling and Autoregressive Models.}
\label{sec:background-ar}
Language models approximate the data distribution $p_\text{data}: \mathcal{V}^L \rightarrow [0, 1]$ over sequences with a density $p_\theta(\x)$. AR models factorize $p_\theta(\x) = \prod_{\ell=1}^L p_\theta(\x^\supl \mid \x^{<\ell})$ using the chain rule of probability. This factorization enables exact likelihood training, but it makes inference slow, since tokens are generated one by one, and it prevents conditioning on future tokens.
\subsection{Flow Generative Modeling}
Continuous Normalizing Flows (CNFs) \citep{Chen2018NeuralOD, grathwohl2018ffjordfreeformcontinuousdynamics} are generative models on $\mathbb{R}^d$. CNFs learn a continuous-time transport from a noise distribution $p_0 = p_\text{noise}$ to the data distribution $p_1 = p_\text{data}$. The time-dependent velocity field $u^\theta_t \colon \mathbb{R}^d \to \mathbb{R}^d$ induces the \emph{flow} $\phi_t \colon \mathbb{R}^d \to \mathbb{R}^d$ by integration:
\begin{equation}
    \label{eq:cnf-ode}
    \frac{d}{dt}\phi_t(z_0) = u^\theta_t(\phi_t(z_0)), \quad \phi_0(z_0) = z_0.
\end{equation}
The velocity field $u^\theta_t$ is parameterized by a neural network with continuously differentiable bounded derivatives, so the ODE \Eqn{eq:cnf-ode} has a unique solution \citep{coddington1955theory}. For $t \in [0, 1]$, the pushforward $p_t = [\phi_t]_\sharp\, p_0$ defines the intermediate density $p_t$, and the family $\{p_t\}_{t \in [0,1]}$ is called a \emph{probability path} from $p_0$ to $p_1$. Integrating \Eqn{eq:cnf-ode} up to $t$ produces a sample $z_t = \phi_t(z_0) \sim p_t$.
\paragraph{Flow Matching.}
Flow Matching (FM) \citep{lipman2023flowmatchinggenerativemodeling, liu2022flowstraightfastlearning, albergo2023buildingnormalizingflowsstochastic} learns the velocity field $u^\theta_t$ that transports $p_0$ to $p_1$ (\supp{supp:background-fm}). The \emph{true} velocity $u_t$ is typically expressed as an expectation:
\begin{equation}
    \label{eq:marginal-vf}
    u_t(z_t) = \int u_{t|1}(z_t \mid x)\, p_{1|t}(x \mid z_t)\, dx,
\end{equation}
where $u_{t|1}$ is a \emph{conditional} velocity given $x \sim p_\text{data}$, and $p_{1|t}$ is the posterior given the noisy example $z_t$. Since $p_{1|t}$ is generally intractable, FM trains $u^\theta_t$ against $u_{t|1}$ instead. Let $\psi_{t|1}$ denote the \emph{conditional} flow associated with $u_{t|1}$. A common choice is the linear interpolation:
\begin{equation}
    \label{eq:conditional-flow}
    \psi_{t|1}(z_0 \mid x) = z_t = \alpha_t\, x + (1 - \alpha_t)\, z_0,
\end{equation}
where $z_0 \sim p_0$ and $x \sim p_1$. The \emph{noise schedule} $\alpha_t \colon [0,1] \to [0,1]$ is monotonically increasing, with $\alpha_0 = 0$ and $\alpha_1 = 1$. The conditional velocity is given by the time derivative of $\psi_{t|1}$:
\begin{equation}
    \label{eq:conditional-vf}
    u_{t|1}(z_t \mid x) = \dot{\alpha}_t\,(x - z_0).
\end{equation}
A key result in FM \citep{lipman2023flowmatchinggenerativemodeling} is that the minimizer of the \emph{Conditional Flow Matching} (CFM) loss is the marginal velocity $u_t$ \Eqn{eq:marginal-vf}:
\begin{equation}
    \label{eq:cfm-loss}
    \mathcal{L}_{\text{CFM}}(\theta) = \mathbb{E}_{t \sim \mathcal{U}[0,1],\, z_0 \sim p_0,\, x \sim p_1} \left\| u^\theta_t(z_t) - u_{t|1}(z_t \mid x) \right\|^2.
\end{equation}
\subsection{Geometry of the Hypersphere}
\label{sec:geom-hypersphere}
We summarize the primitives we use on $\mathbb{S}^{d-1}$. For a thorough treatment of Riemannian geometry, see \citet{docarmo1992riemannian}. The \emph{geodesic distance} between $\mathbf{p}, \mathbf{q} \in \mathbb{S}^{d-1}$ is the angle $d_{\mathbb{S}}(\mathbf{p}, \mathbf{q}) = \arccos(\mathbf{p}^\top \mathbf{q}) \in [0, \pi]$. The \emph{tangent space} at $\mathbf{p}$ is $T_{\mathbf{p}} \mathbb{S}^{d-1} = \{\mathbf{v} \in \mathbb{R}^d : \mathbf{v}^\top \mathbf{p} = 0\}$. The \emph{exponential map} $\exp_{\mathbf{p}} \colon T_{\mathbf{p}} \mathbb{S}^{d-1} \to \mathbb{S}^{d-1}$ moves $\mathbf{p}$ along the geodesic in direction $\mathbf{v}$ for arc length $\|\mathbf{v}\|$:
\begin{equation}
    \label{eq:exp-map}
    \exp_{\mathbf{p}}(\mathbf{v}) = \cos(\|\mathbf{v}\|)\,\mathbf{p} + \sin(\|\mathbf{v}\|)\,\frac{\mathbf{v}}{\|\mathbf{v}\|}.
\end{equation}
The \emph{logarithmic map} $\log_{\mathbf{p}} \colon \mathbb{S}^{d-1} \to T_{\mathbf{p}} \mathbb{S}^{d-1}$ inverts the exponential map and returns the tangent vector at $\mathbf{p}$ pointing toward $\mathbf{q}$, with magnitude $d_{\mathbb{S}}(\mathbf{p}, \mathbf{q})$\footnote{When $\mathbf{p} = -\mathbf{q}$ (antipodal points), $\log_{\mathbf{p}}(\mathbf{q})$ is undefined since infinitely many geodesics connect them.}:
\begin{equation}
    \label{eq:log-map}
    \log_{\mathbf{p}}(\mathbf{q}) = \frac{\omega}{\sin \omega}\left(\mathbf{q} - \cos(\omega)\,\mathbf{p}\right), \quad \omega = d_{\mathbb{S}}(\mathbf{p}, \mathbf{q}).
\end{equation}
The \emph{Spherical Linear Interpolation} (SLERP) follows the geodesic from $\mathbf{p}$ to $\mathbf{q}$ \citep{SLERP}:
\begin{equation}
    \label{eq:slerp}
    \mathrm{SLERP}(\mathbf{p}, \mathbf{q}, t) = \frac{\sin((1{-}t)\,\omega)}{\sin \omega}\,\mathbf{p} + \frac{\sin(t\,\omega)}{\sin \omega}\,\mathbf{q}, \quad \omega = d_{\mathbb{S}}(\mathbf{p}, \mathbf{q}),
\end{equation}
or equivalently, $\mathrm{SLERP}(\mathbf{p}, \mathbf{q}, t) = \exp_{\mathbf{p}}\!\left(t\,\log_{\mathbf{p}}(\mathbf{q})\right)$. To sample uniformly on $\mathbb{S}^{d-1}$, draw \mbox{$\boldsymbol{\epsilon} \sim \mathcal{N}(\mathbf{0}, \mathbf{I}_d)$} and normalize $\boldsymbol{\epsilon} \leftarrow \boldsymbol{\epsilon} / \|\boldsymbol{\epsilon}\|$ \citep{muller1959note}.
\subsection{Flow Matching on the Hypersphere}
\label{sec:fm-on-hypersphere}
Riemannian Flow Matching (RFM) \citep{chen2024flowmatchinggeneralgeometries} extends FM to Riemannian manifolds, such as $\mathbb{S}^{d-1}$. As in the Euclidean case, the marginal velocity is the expectation of a conditional velocity. In particular, the following posterior-averaged velocity field satisfies the continuity equation on the manifold for the induced marginal probability path (\supp{sec:hyperspherical-continuity-equation}):
\begin{equation}
    \label{eq:riemannian-marginal-vf}
    u_t(\mathbf{z}_t) = \int u_{t|1}(\mathbf{z}_t \mid \mathbf{z}_1)\, p_{1|t}(\mathbf{z}_1 \mid \mathbf{z}_t)\, d\mathbf{z}_1,
\end{equation}
where $u_{t|1}$ is the conditional velocity associated with the conditional flow $\psi_{t|1}$, and $p_{1|t}$ is the posterior given $\mathbf{z}_t$. We define $\psi_{t|1}$ with SLERP \Eqn{eq:slerp}, where $\mathbf{z}_0 \sim p_0$ is drawn from a noise distribution on $\mathbb{S}^{d-1}$ and $\mathbf{z}_1 \sim p_1$ is a data sample:
\begin{equation}
    \label{eq:spherical-conditional-flow}
    \psi_{t|1}(\mathbf{z}_0 \mid \mathbf{z}_1) = \mathbf{z}_t = \mathrm{SLERP}(\mathbf{z}_0, \mathbf{z}_1, \alpha_t) = \exp_{\mathbf{z}_0}\!\left(\alpha_t\,\log_{\mathbf{z}_0}(\mathbf{z}_1)\right),
\end{equation}
which satisfies $\psi_{0|1}(\mathbf{z}_0 \mid \mathbf{z}_1) = \mathbf{z}_0$ and $\psi_{1|1}(\mathbf{z}_0 \mid \mathbf{z}_1) = \mathbf{z}_1$. Differentiating $\psi_{t|1}$ gives the conditional velocity (\supp{sec:vf-derivation}):
\begin{equation}
    \label{eq:spherical-conditional-vf}
    u_{t|1}(\mathbf{z}_t \mid \mathbf{z}_1) = \frac{\dot{\alpha}_t}{1 - \alpha_t}\,\log_{\mathbf{z}_t}(\mathbf{z}_1).
\end{equation}
As in the Euclidean case, the minimizer of the Riemannian Conditional Flow Matching (RCFM) loss is the marginal velocity $u_t$ \Eqn{eq:riemannian-marginal-vf}:
\begin{equation}
    \label{eq:rcfm-loss}
    \mathcal{L}_{\text{RCFM}}(\theta) = \mathbb{E}_{t \sim \mathcal{U}[0,1],\, \mathbf{z}_0 \sim p_0,\, \mathbf{z}_1 \sim p_1} \left\| u^\theta_t(\mathbf{z}_t) - u_{t|1}(\mathbf{z}_t \mid \mathbf{z}_1) \right\|^2.
\end{equation}
\section{Hyperspherical Flow Language Models}
\label{sec:method}
\Fig{fig:S-FLM-fig1} gives an overview of training and sampling. \method is a Riemannian CNF on $(\mathbb{S}^{d-1})^L$ that transports a sequence of random vectors on $\mathbb{S}^{d-1}$ toward the clean token representations. To bridge the continuous and discrete representations, we map tokens to the sphere via a normalized embedding lookup and decode via the $\arg\max$ of $p^\theta_{1|t}$. The denoiser $p^\theta_{1|t}$, trained with Cross-Entropy, induces a closed-form marginal velocity field that we integrate at sampling time. Optionally, after each optimization step, we re-project the embeddings to $\mathbb{S}^{d-1}$ (\supp{sec:forced-normalization}).
\paragraph{Encoder and Decoder.}
Let $\mathbf{x} = (x^1, \ldots, x^L) \in \mathcal{V}^L$ be an input sequence of $L$ tokens. Each token $v \in \mathcal{V}$ is associated with a unit-norm embedding $\hat{\mathbf{e}}_v \in \mathbb{S}^{d-1}$, so we represent $\mathbf{x}$ as a sequence in $(\mathbb{S}^{d-1})^L$. The decoder $g \colon (\mathbb{S}^{d-1})^L \times [0,1] \to \mathcal{V}^L$ is defined as the $\arg\max$ of the learned posterior:
\begin{equation}
    \label{eq:encoder-decoder}
    \mathbf{e}_v = \mathbf{E}[v], \qquad \hat{\mathbf{e}}_v = \frac{\mathbf{e}_v}{\|\mathbf{e}_v\|_2}, \qquad g^\supl(\mathbf{z}, t) = \arg\max_{v \in \mathcal{V}} \, p^\theta_{1|t}(\x^\supl = v \mid \mathbf{z}).
\end{equation}
\paragraph{Training with Cross-Entropy.}
Unlike standard flow matching, which operates on fixed data (e.g., pixels), \method learns the data representation, an embedding table $\mathbf{E}$, jointly with the flow by backpropagating through the SLERP. Regressing the velocity field against learnable embeddings admits a trivial minimum where all token representations collapse to a point. Instead, we approximate the posterior $p_{1|t}(\cdot \mid \mathbf{z}_t)$ with a denoiser $p^\theta_{1|t}(\cdot \mid \mathbf{z}_t)$, trained with Cross-Entropy. Because the denoiser cannot recover the clean token if embeddings collapse, the CE loss pushes them apart, as noted in CDCD \citep{dieleman2022continuousdiffusioncategoricaldata}. At every position $\ell$, we take $\mathbf{z}_1^\supl = \hat{\mathbf{e}}_{\x^\supl}$ \Eqn{eq:encoder-decoder} as the clean endpoint, draw $\mathbf{z}_0^\supl \sim \mathcal{U}(\mathbb{S}^{d-1})$ independently, and form the noisy latent $\mathbf{z}_t^\supl = \mathrm{SLERP}(\mathbf{z}_0^\supl, \mathbf{z}_1^\supl, \alpha_t)$ \Eqn{eq:spherical-conditional-flow}. We minimize the Cross-Entropy
\begin{equation}
    \label{eq:ce-loss}
    \mathcal{L}_{\text{CE}}(\theta) = \mathbb{E}_{\mathbf{x} \sim p_1,\, t \sim \mathcal{U}[0,1],\, \mathbf{z}_0 \sim p_0} \left[ -\sum_{\ell=1}^L \log p^\theta_{1|t}(\x^\supl \mid \mathbf{z}_t) \right].
\end{equation}
\subsection{Sampling}
\paragraph{Exact Velocity.}
\label{par:exact-velocity}
Because our data distribution is supported on $\{\hat{\mathbf{e}}_v : v \in \mathcal{V}\}^L \subset (\mathbb{S}^{d-1})^L$, the marginal velocity \Eqn{eq:riemannian-marginal-vf} reduces to a finite sum at each position:
\begin{equation}
    \label{eq:flm-marginal-vf}
    u^\theta_t(\mathbf{z}_t^\supl) = \frac{\dot{\alpha}_t}{1 - \alpha_t} \sum_{v \in \mathcal{V}} p^\theta_{1|t}(\x^\supl = v \mid \mathbf{z}_t)\, \log_{\mathbf{z}_t^\supl}\!\left(\hat{\mathbf{e}}_v\right).
\end{equation}
We call this the \emph{exact velocity}, in contrast to the stochastic and top-$k$ approximations below.
\paragraph{Top-$k$ Velocity.}
\label{par:top-k-velocity}
The exact velocity is Bayes-optimal when the learned posterior is calibrated. In practice, when the posterior remains diffuse over semantically incompatible continuations, posterior averaging can move the latent toward a region that corresponds to no valid token-level mode. Top-$k$ decoding replaces the averaging in \Eqn{eq:flm-marginal-vf} by truncating the posterior $p^\theta_{1|t}(\cdot \mid \mathbf{z}_t)$ to its $k$ largest entries. In particular, top-$1$ decoding behaves like a greedy mode-following approximation. Top-$1$ decoding improves accuracy on verifiable reasoning but departs from the exact marginal flow.
\paragraph{Stochastic Decoding.}
\label{par:stochastic-decoding}
We can replace the sum in \Eqn{eq:flm-marginal-vf} with a single Monte Carlo sample of the posterior. At each step, we draw $\hat{\x}^\supl \sim p^\theta_{1|t}(\cdot \mid \mathbf{z}_t)$ and use $\bar{\mathbf{v}}^\supl = \log_{\mathbf{z}_t^\supl}(\hat{\mathbf{e}}_{\hat{\x}^\supl})$ in place of \Eqn{eq:flm-marginal-vf}. The deterministic and stochastic samplers differ only in how they construct the velocity from $p^\theta_{1|t}$. CANDI \citep{pynadath2025candihybriddiscretecontinuousdiffusion} uses an analogous one-sample Monte Carlo approximation. See \algo{alg:sampling-stochastic} for the pseudocode.
\paragraph{Temperature Scaling.}
\label{par:temperature-scaling}
The three velocity variants depend on the posterior $p^\theta_{1|t}(\cdot \mid \mathbf{z}_t)$. To control sampling diversity, we apply a temperature $T > 0$ to the logits $\zeta_v$ of $p^\theta_{1|t}$:
\begin{equation}
    \label{eq:temperature-scaling}
    p^{\theta,T}_{1|t}(v \mid \mathbf{z}_t) \propto \exp(\zeta_v / T).
\end{equation}
When sampling with scaled temperature, we substitute $p^{\theta,T}_{1|t}$ for $p^\theta_{1|t}$ in the exact, stochastic, and top-$k$ velocities.
\subsection{Noise Schedule}
\paragraph{Truncation.}
After training with standard noise schedules (\tab{tab:noise-schedules}), we observe that $p^\theta_{1|t}$ becomes close to one-hot after only a few sampling steps. This is likely due to the \emph{curse of dimensionality} \citep{bellman1961adaptive}. In high dimension, $\mathbb{S}^{d-1}$ has enough room to spread $|\mathcal{V}|$ embeddings apart \citep{vershynin2018high}. Training with CE separates the embeddings because the denoiser cannot distinguish tokens whose embeddings coincide \citep{wang2020understandingcontrastiverepresentationlearning}. When the embeddings are well separated, the true posterior $p_{1|t}$ at low noise levels ($\alpha_t$ large) is close to one-hot. Thus, during sampling, after $\mathbf{z}_t^\supl$ enters the Voronoi cell of a clean embedding $\hat{\mathbf{e}}_j$, the posterior collapses. To avoid training on noise levels where $p_{1|t}$ is close to one-hot, we truncate the noise schedule to $[0, a] \subseteq [0, 1]$, i.e., we train only at the higher noise levels. By analyzing a tractable approximation of the sampling dynamics on $\mathbb{S}^{d-1}$, we derive a closed-form expression for the truncation bound $a = \alpha^\star(\delta)$ as a function of the vocabulary size $|\mathcal{V}|$ and the embedding dimension $d$. Truncation is critical for strong performance on Sudoku (\tab{tab:sudoku}), and on GSM8K our bound achieves accuracy similar to that of the best truncation value found by grid search (\supp{sec:appendix:tinygsm-ablations}).
\begin{tcolorbox}[
  colback=ourlightblue!12!white,
  colframe=ourlightblue!60!ourblue!80!black,
  boxrule=1pt,
  arc=3pt,
  left=5pt, right=5pt,
  top=3pt, bottom=3pt,
  boxsep=2pt,
  before skip=4pt, after skip=4pt
]
\noindent\textbf{Tractable model of the sampling dynamics.}
Let $\{\hat{\mathbf{e}}_v\}_{v \in \mathcal{V}}$ be the normalized token representations, sampled i.i.d.\ uniformly on $\mathbb{S}^{d-1}$. Let $\mathbf{z}_0 \sim \mathcal{U}(\mathbb{S}^{d-1})$ be the initial noise sample. Fix a target token $j$ and let $\mathbf{z}_\alpha = \mathrm{SLERP}(\mathbf{z}_0, \hat{\mathbf{e}}_j, \alpha)$ for $\alpha \in [0,1]$. Define $\alpha^\star(\delta)$ as the smallest $\alpha$ at which $\hat{\mathbf{e}}_j$ is the nearest neighbor of $\mathbf{z}_\alpha$ with probability at least $1-\delta$. Then, in high dimension,
\begin{equation}
    \label{eq:alpha-crit}
    \alpha^\star(\delta)
    \approx
    \frac{2}{\pi}
    \arcsin\!\left(
        \sqrt{\frac{2\log\!\bigl(2(|\mathcal{V}|-1)/\delta\bigr)}{d}}
    \right).
\end{equation}
\end{tcolorbox}
\begin{proof}[Derivation sketch]
Under our model, the similarity to non-target tokens $\langle \mathbf{z}_\alpha, \hat{\mathbf{e}}_v \rangle$ (with $v \neq j$) is sub-Gaussian, as the inner product of a fixed vector and a uniform random vector on $\mathbb{S}^{d-1}$ \citep{vershynin2018high}. With a simple union bound, we conclude that $\max_{v \neq j} \langle \mathbf{z}_\alpha, \hat{\mathbf{e}}_v \rangle \le \sqrt{2\log(2(|\mathcal{V}|-1)/\delta)/d}$ with probability at least $1-\delta$. Along the sampling trajectory, $\langle \mathbf{z}_\alpha, \hat{\mathbf{e}}_j \rangle = \cos((1-\alpha)\omega)$ with $\omega = d_{\mathbb{S}}(\mathbf{z}_0, \hat{\mathbf{e}}_j) \approx \pi/2$ in high dimension, so $\langle \mathbf{z}_\alpha, \hat{\mathbf{e}}_j \rangle \approx \sin(\pi\alpha/2)$. We conclude by solving for the critical point such that $\mathbf{z}_\alpha$ is closest to $\hat{\mathbf{e}}_j$ with high probability. The complete argument is in \supp{sec:formal-random-codebook-alpha-crit}, and the numerical values in \tab{tab:alpha-crit-grid}.
\end{proof}
\paragraph{Adaptive Noise Schedule.}
Inspired by InfoNoise \citep{raya2026informationguidednoiseallocation}, we also adapt the truncated noise schedule during training to allocate more samples to noise levels where the loss $\mathcal{L}$ changes most. Every 50 steps, we fit the loss profile $\hat{\mathcal{L}}(t)$ from recent pairs $(t, \mathcal{L})$ and define the noise schedule $\alpha_t$ using the inverse CDF of $|d\hat{\mathcal{L}}/dt|$ \citep{dieleman2022continuousdiffusioncategoricaldata}. We use $|d\hat{\mathcal{L}}/dt|$ as a proxy for the critical noise levels where the model learns most, and thus sample these levels more often. We stabilize the updates with an \emph{Exponential Moving Average} (EMA) of the successive
schedules. See \supp{sec:adaptive-schedule} for the pseudocode and a comparison with InfoNoise. In practice, the adaptive schedule does not slow training (\tab{tab:training-cost}).
\subsection{Hyperspherical Architecture}
\label{sec:hyperspherical-architecture}
The latents $\mathbf{z}_t^\supl$ live on $\mathbb{S}^{d-1}$, and the sampling step \Eqn{eq:flm-marginal-vf} rotates vectors. We propose a Transformer variant inspired by nGPT \citep{loshchilov2024ngptnormalizedtransformerrepresentation} that keeps the intermediate activations on $\mathbb{S}^{d-1}$ and parameterizes the attention and MLP layers as rotations. Each block replaces the additive residual with a normalized interpolation $\mathbf{h}_{\mathrm{out}} \leftarrow \mathrm{Norm}\bigl(\mathbf{h}_{\mathrm{in}} + \boldsymbol{\gamma} \odot (\mathrm{Norm}(\mathbf{h}_{\mathrm{layer}}) - \mathbf{h}_{\mathrm{in}})\bigr)$, where $\mathbf{h}_{\mathrm{in}}$ is the input, $\mathbf{h}_{\mathrm{layer}}$ is the layer (MLP or attention) output, and $\mathbf{h}_{\mathrm{out}}$ is the updated state. The normalized interpolation approximates SLERP for small $\boldsymbol{\gamma}$. The per-dimension gate $\boldsymbol{\gamma}$ is computed from the noise level, similar to Adaptive Layer Norm (adaLN) in the \emph{Diffusion Transformer} (DiT) \citep{peebles2023scalablediffusionmodelstransformers}. We do not use adaLN, so our architecture (\emph{\arch}) has slightly \emph{fewer} parameters than the standard DiT used in discrete diffusion papers \citep{sahoo2024simpleeffectivemaskeddiffusion, sahoo2025diffusionduality, sahoo2026scalingmaskeddiffusionlanguage}. The \arch achieves better results than the standard DiT (\sec{sec:experiments}).
\begin{tcolorbox}[
  colback=ourlightblue!12!white,
  colframe=ourlightblue!60!ourblue!80!black,
  boxrule=1pt,
  arc=3pt,
  left=5pt, right=5pt,
  top=3pt, bottom=3pt,
  boxsep=2pt,
  before skip=4pt, after skip=4pt
]
\textbf{Takeaway.} We propose the \arch to implement the denoiser $p^\theta_{1|t}$ in place of the standard DiT. We train with Cross-Entropy \Eqn{eq:ce-loss} under a truncated, adaptive noise schedule. We marginalize the conditional velocities under $p^\theta_{1|t}$ to obtain $u^\theta_t$ \Eqn{eq:flm-marginal-vf}, integrate it from $\mathbf{z}_0 \sim p_0$ to $\mathbf{z}_1$, and decode with $\arg\max$. The velocity admits exact, stochastic, and top-$k$ variants. See \algo{alg:training}, \algo{alg:sampling}, and \algo{alg:sampling-stochastic} for pseudocode.
\end{tcolorbox}
\begin{figure}[t]
    \centering
    \begin{subfigure}[b]{0.48\textwidth}
        \centering
        \includegraphics[width=\linewidth]{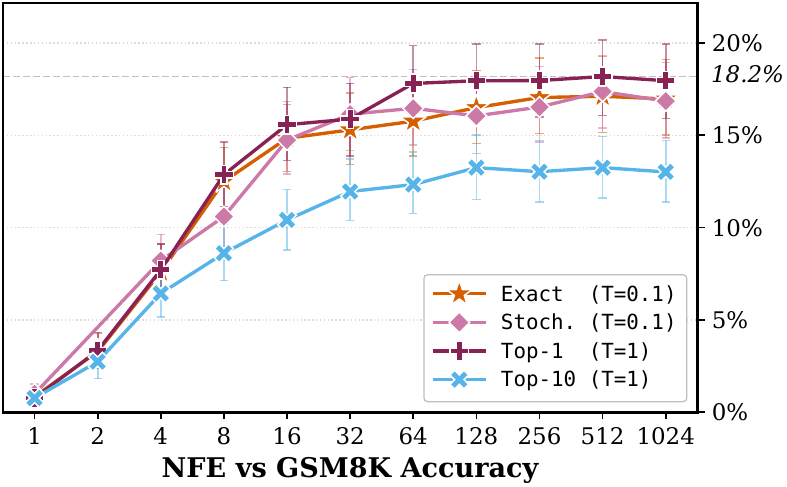}
        \label{fig:tinygsm-velocity-topk-T0p1}
    \end{subfigure}\hfill
    \begin{subfigure}[b]{0.48\textwidth}
        \centering
        \includegraphics[width=\linewidth]{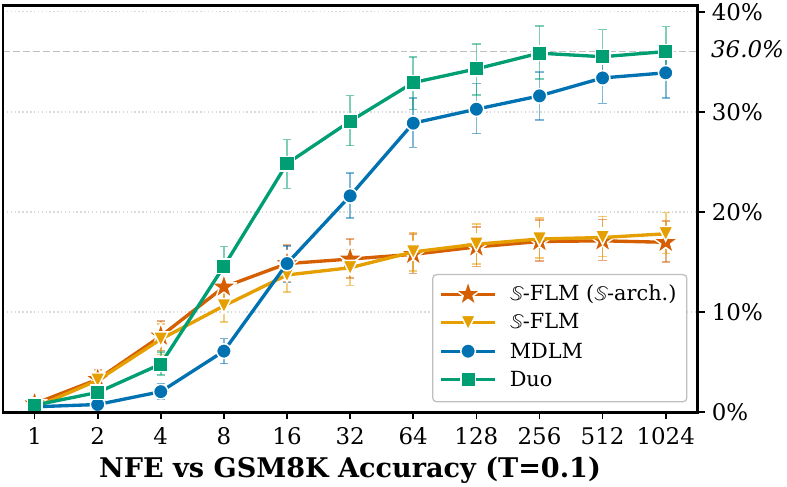}
        \label{fig:tinygsm-arch-baselines-T0p1}
    \end{subfigure}
    \caption{\textbf{Accuracy on GSM8K at $T = 0.1$.} \textbf{Left:} Decoding strategies for \method (\arch). At low temperature, sampling with the exact or stochastic velocities approaches the accuracy of top-$1$ decoding. \textbf{Right:} At $T=0.1$ the standard DiT and the \arch perform similarly, and their accuracy is roughly half that of Duo. At $T = 1$ the \arch outperforms the standard DiT (\Fig{fig:appendix-tinygsm-arch-baselines-T1p0}).}
    \label{fig:tinygsm-T0p1}
\end{figure}
\section{Experiments}
\label{sec:experiments}
We apply \method to Sudoku solving (\sec{sec:experiments:sudoku}), math reasoning via code on TinyGSM \citep{liu2023tinygsm} (\sec{sec:experiments:tinygsm}), and unconditional language modeling on OWT \citep{Gokaslan2019OpenWeb} (\sec{sec:experiments:owt}). In recent work on diffusion language models, the most common measure of sample quality is Gen.\ PPL, computed with a large AR model. However, samples with good likelihood are not necessarily correct at the sequence level \citep{velickovic2026perplexitycannotalwaystellright, feng2025theoreticalbenefitlimitationdiffusion}, and repetitive text has low perplexity \citep{dieleman2022continuousdiffusioncategoricaldata, deschenaux2024promisesoutlookschallengesdiffusion}. Therefore, we primarily focus on datasets with ground-truth solutions (Sudoku, GSM8K).
\newpage
\subsection{Reasoning on Sudoku}
\label{sec:experiments:sudoku}
\begin{wraptable}{r}{0.5\textwidth}
    \vspace{-1.4em}
    \caption{Exact-match accuracy (\%) on Sudoku when sampling with 180 steps. The overall best is \underline{underlined}. The best score with continuous diffusion is \textbf{bolded}. CANDI is a hybrid continuous-masked model, which can also be trained as a pure Gaussian diffusion model. \method is the best continuous model, closely followed by the FLM baseline.}
    \label{tab:sudoku}
    \centering
    \small
    \setlength{\tabcolsep}{6pt}
    \renewcommand{\arraystretch}{1.05}
        {%
        \newcommand{\tabrow}{\hspace*{0.8em}}
        \begin{tabular}{l ccc}
        \toprule
        Model & Easy & Med. & Hard \\
        \midrule
        \multicolumn{4}{@{}l@{}}{\textit{Autoregressive}} \\
        \tabrow Sample & 13.9 & 5.1 & 0.6 \\
        \tabrow Greedy & 14.6 & 5.1 & 1.0 \\
        \midrule
        \multicolumn{4}{@{}l@{}}{\textit{Discrete Diffusion}} \\
        \tabrow MDLM {\tiny\citep{sahoo2024simpleeffectivemaskeddiffusion}} & 92.0 & 77.1 & 30.2 \\
        \tabrow Duo {\tiny\citep{sahoo2025diffusionduality}} & \underline{96.3} & 84.7 & \underline{58.4} \\
        \midrule
        \multicolumn{4}{@{}l@{}}{\textit{Continuous Diffusion}} \\
        \tabrow CANDI {\tiny\citep{pynadath2025candihybriddiscretecontinuousdiffusion}} & 79.3 & 45.9 & 16.7 \\
        \tabrow CANDI {\tiny pure Gaussian} & 63.9 & 41.9 & 12.5 \\
        \tabrow FLM {\tiny\citep{lee2026onesteplanguagemodelingcontinuous}} & 94.2 & 82.7 & 44.5 \\
        \rowcolor{gray!15}
        \tabrow \method & 81.5 & 50.6 & 14.0 \\
        \rowcolor{gray!15}
        \tabrow \method {\tiny + $\alpha^\star(0.1)$} & 94.0 & 77.6 & 43.2 \\
        \rowcolor{gray!15}
        \tabrow \method {\tiny + $\alpha^\star(0.1)$ + adaptive} & \textbf{94.8} & \underline{\textbf{85.2}} & \textbf{45.0} \\
        \bottomrule
        \end{tabular}%
    }
    \vspace{-2.5em}
\end{wraptable}
\paragraph{Experimental Setup.}
We evaluate \method on Sudoku \citep{benhamu2025acceleratedsamplingmaskeddiffusion, kim2025finetuningmaskeddiffusionprovable}. We use 48k training and 2k validation puzzles (no overlap), each with a unique solution \citep{alp2024sudoku}. We define three difficulty levels based on the number of visible digits (easy: 40/81, medium: 35/81, hard: 30/81). We use the modified DiT \citep{peebles2023scalablediffusionmodelstransformers} architecture from SEDD \citep{lou2024discretediffusionmodelingestimating} with 8 layers and embedding dimension 512. We compare against AR, MDLM \citep{sahoo2024simpleeffectivemaskeddiffusion}, Duo \citep{sahoo2025diffusionduality}, CANDI \citep{pynadath2025candihybriddiscretecontinuousdiffusion}, and FLM \citep{lee2026onesteplanguagemodelingcontinuous}, using 180 sampling steps for the diffusion variants. We describe the input format and training hyperparameters in \supp{sec:appendix:sudoku-setup}.
\paragraph{Results.}
\tab{tab:sudoku} shows the results. The autoregressive model performs poorly at all difficulty levels, since the task requires global context. Duo \citep{sahoo2025diffusionduality} obtains the highest accuracy on easy and hard puzzles. Among the continuous methods, FLM \citep{lee2026onesteplanguagemodelingcontinuous} and \method outperform AR and MDLM at every difficulty. \method with the simple linear schedule performs poorly on hard puzzles, but with truncation and the adaptive schedule, \method performs similarly to the best prior continuous model (the FLM baseline). See \supp{sec:appendix:sudoku-ablations} for the ablation over schedules and embedding re-projection. We do not re-project embeddings to $\mathbb{S}^{d-1}$ after each optimization step in \tab{tab:sudoku}.
\begin{figure}[!b]
    \centering
    \begin{subfigure}[t]{0.49\textwidth}
        \centering
        \includegraphics[width=\linewidth]{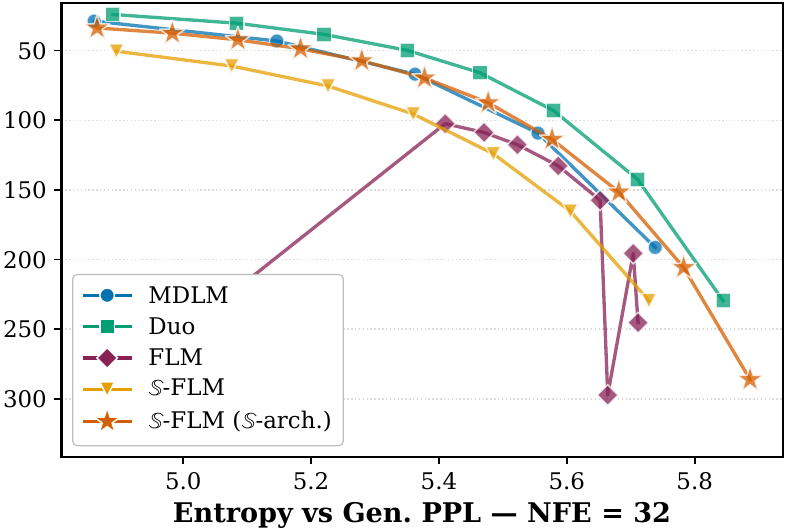}
    \end{subfigure}\hfill
    \begin{subfigure}[t]{0.49\textwidth}
        \centering
        \includegraphics[width=\linewidth]{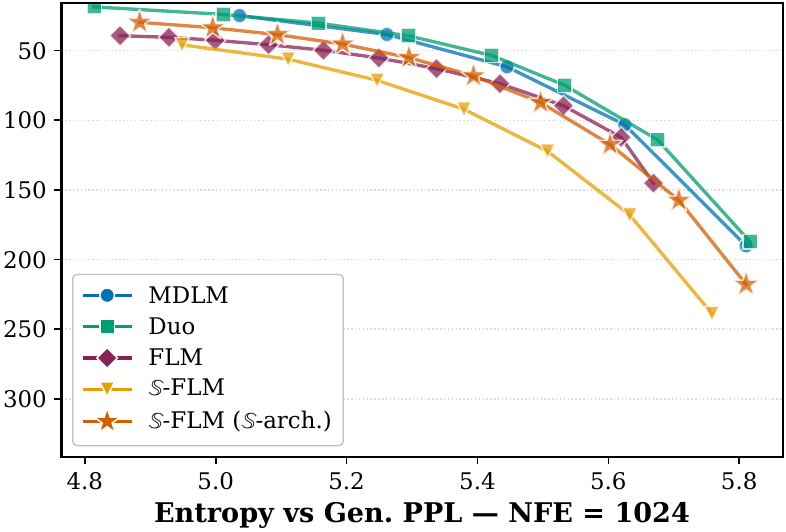}
    \end{subfigure}
    \caption{\textbf{Gen.\ PPL ($\downarrow$) / Entropy ($\uparrow$) Frontier on OpenWebText} at NFE $= 32$ (left) and NFE $= 1024$ (right). Each curve sweeps the temperature $T$. \method with the \arch performs similarly to prior FLMs. Duo is best overall. At NFE $= 32$, the frontier of the FLM baseline is highly unstable.}
    \label{fig:owt-pareto-main}
\end{figure}
\newpage
\subsection{Reasoning on GSM8K}
\label{sec:experiments:tinygsm}
\begin{wraptable}{l}{0.38\textwidth}
    \vspace{-1.5em}
    \caption{Accuracy (\%) on GSM8K after 250k training steps. Diffusion variants use 1024 steps and no temperature scaling, and AR uses 512 (the context length). The overall best is \underline{underlined} and the best continuous diffusion result is \textbf{bolded}. \method outperforms prior continuous diffusion language models, and \textbf{closes the gap to MDLM at $T = 1$ when the velocity \Eqn{eq:flm-marginal-vf} is computed with top-$1$.} A gap remains at low temperature ($T = 0.1$, \Fig{fig:tinygsm-T0p1}).}
    \label{tab:tinygsm_main}
    \centering
    \small
    \vspace{-0.6em}
    {
    \newcommand{\tabrow}{\hspace*{0.8em}}
    \begin{tabular}{lr}
    \toprule
    Model & Acc.\ (\%) \\
    \midrule
    \multicolumn{2}{@{}l@{}}{\textit{Autoregressive}} \\
    \tabrow Sample & 53.9 \\
    \tabrow Greedy & \underline{63.3} \\
    \midrule
    \multicolumn{2}{@{}l@{}}{\textit{Discrete Diffusion}} \\
    \tabrow MDLM {\tiny\citep{sahoo2024simpleeffectivemaskeddiffusion}} & 18.0 \\
    \tabrow Duo {\tiny\citep{sahoo2025diffusionduality}} & 17.2 \\
    \midrule
    \multicolumn{2}{@{}l@{}}{\textit{Continuous Diffusion}} \\
    \tabrow CANDI {\tiny\citep{pynadath2025candihybriddiscretecontinuousdiffusion}} & 0.2 \\
    \tabrow FLM {\tiny\citep{lee2026onesteplanguagemodelingcontinuous}} & 0.3 \\
    \rowcolor{gray!15}
    \tabrow \method & 1.2 \\
    \tabrow \quad + $\alpha^\star(0.1)$ & 7.7 \\
    \tabrow \quad + no renorm. & 8.1 \\
    \tabrow \quad + adaptive & 11.1 \\
    \tabrow \quad + \arch. {\small (\sec{sec:hyperspherical-architecture})} & 12.4 \\
    \tabrow \quad + top-$1$ & \textbf{18.0} \\
    \bottomrule
    \end{tabular}%
    }
    \vspace{-2.4em}
\end{wraptable}
\paragraph{Experimental Setup.}
TinyGSM \citep{liu2023tinygsm} is a dataset of $\sim$11.8M synthetic math word problems similar to GSM8K \citep{cobbe2021trainingverifierssolvemath}. Each solution is a GPT-3.5-generated Python program that produces the numerical answer. We evaluate all models on the GSM8K test set by executing one generated solution per problem. Error bars show bootstrapped $95\%$ confidence intervals over the test set (\supp{sec:appendix:bootstrap}). We use the \texttt{SmolLM-135M} tokenizer \citep{allal2025smollm2smolgoesbig} because it compresses code better than the GPT-2 tokenizer \citep{Radford2019LanguageMA} (\Fig{fig:tokenlen-smollm-gpt2}). We pad the input sequences to length 512 and compute the loss on padding tokens. As with Sudoku, we always keep the problem statement clean so that we can sample conditionally. All backbones use hidden dimension 768, 12 layers, 12 attention heads, and dropout 0.1. For \method, we train both the standard DiT and the \arch (\sec{sec:hyperspherical-architecture}).
We train for 250k steps with batch size 512, using Adam ($\text{lr} = 3 \times 10^{-4}$, $\beta_1 = 0.9$, $\beta_2 = 0.999$, no weight decay) and an EMA rate of $0.9999$.
\paragraph{Results.}We report the accuracy at $T=1$ with 1024 sampling steps for diffusion variants in \tab{tab:tinygsm_main}. CANDI \citep{pynadath2025candihybriddiscretecontinuousdiffusion} and FLM \citep{lee2026onesteplanguagemodelingcontinuous} reach less than $1\%$ accuracy, even with their respective optimized schedules, and temperature annealing does not improve their accuracy beyond $0.5\%$. In contrast, \method solves $18\%$ of problems with top-$1$ decoding. \tab{tab:tinygsm_main} also ablates each new component of \method. With the vanilla linear schedule, \method reaches only $1.2\%$.
Truncating the noise schedule according to \Eqn{eq:alpha-crit} improves the accuracy to 7.7\%. \emph{Not} re-projecting the embeddings after each optimization step (\supp{sec:forced-normalization}) increases it to 8.1\%. The denoiser always processes unit-norm vectors in the forward pass, but the embedding table itself can either be re-projected to $\mathbb{S}^{d-1}$ after every optimization step or left unconstrained. Skipping the re-projection is equivalent to annealing the learning rate of the embedding vectors, which might stabilize training since the velocity \Eqn{eq:flm-marginal-vf} is a function of the embeddings. The adaptive noise schedule increases the accuracy to 11.1\%, and the \arch increases it further to 12.4\%. With top-$1$ velocity decoding \Eqn{eq:flm-marginal-vf}, \method closes the gap to MDLM and Duo at $T = 1$.
\paragraph{Ablations.}
\Fig{fig:tinygsm-T1p0} shows that at $T = 1$ and NFE $\leq 16$, \method with the exact velocity outperforms MDLM and Duo. All methods benefit from greedy decoding (\sec{par:top-k-velocity}), so we compare \method with MDLM and Duo at low temperature ($T = 0.1$). At $T = 0.1$, MDLM reaches $33\%$ and Duo $36\%$, while \method reaches about $18\%$ (\Fig{fig:tinygsm-T0p1}). Thus, although \method outperforms previous continuous approaches, it remains well behind discrete diffusion and AR models at low temperature. \Fig{fig:appendix-tinygsm-velocity-topk-norm-T1p0} shows that at $T = 1$, top-$k$ decoding significantly improves the accuracy only for $k = 1$. In addition, top-$1$ decoding at $T = 1$ beats the stochastic sampler at low temperature (\Fig{fig:appendix-tinygsm-velocity-topk-T0p1-vs-T1p0}). See \supp{sec:appendix:tinygsm-additional} for more details.
\subsection{Language Modeling on OpenWebText}
\label{sec:experiments:owt}
\paragraph{Experimental Setup.}Following MDLM \citep{sahoo2024simpleeffectivemaskeddiffusion}, we train \method on OpenWebText using the GPT-2 tokenizer and a context length of 1024. We train both the standard DiT and the \arch, each with 12 layers, hidden dimension 768, and 12 attention heads. We train for 1M steps with the same optimizer configuration as in TinyGSM (\sec{sec:experiments:tinygsm}). For FLM \citep{lee2026onesteplanguagemodelingcontinuous}, we use the authors' original checkpoint. %
\paragraph{Gen.\ PPL / Entropy Frontier.}Repetitive generations can improve the Gen.\ PPL without improving sample quality \citep{velickovic2026perplexitycannotalwaystellright}, and different models might have different optimal sampling temperatures. We therefore evaluate the Gen.\ PPL and the average unigram entropy across $T \in \{0.70, 0.75, \ldots, 1.20\}$ and plot the frontier for each model and NFE following CANDI \citep{pynadath2025candihybriddiscretecontinuousdiffusion} (details in \supp{sec:appendix:owt-protocol}). At NFE $= 1024$ (right), \method with the \arch matches prior FLMs in Gen.\ PPL, while the standard DiT is slightly weaker. At low NFE, the frontier becomes unstable for prior FLMs, but remains stable for \method, reaching a Gen.\ PPL similar to MDLM's. Duo generally achieves the best Gen.\ PPL / Entropy trade-off. See \supp{sec:appendix:owt-additional} for the complete sweep of NFEs.
\section{Related Work}
\label{sec:related-work}
\method differs from prior work in three ways: it operates in continuous rather than discrete space, defines its flow on the hypersphere $\mathbb{S}^{d-1}$, and learns token embeddings end-to-end.
\paragraph{Continuous Diffusion for Language Modeling.}Prior work applies Gaussian diffusion to embeddings and trains end-to-end with Cross-Entropy \citep{li2022diffusionlmimprovescontrollabletext, gulrajani2023likelihoodbaseddiffusionlanguagemodels}, or regresses onto pre-trained embeddings \citep{strudel2022selfconditionedembeddingdiffusion, lovelace2022latentdiffusionlanguagegeneration, shen2026codarcontinuousdiffusionlanguage}, which caps sample quality at the pre-trained embeddings and requires two training stages. Riemannian flow models extend score-based generative modeling to manifolds \citep{mathieu2020riemanniancnf, debortoli2022riemannianscorebased, lou2023scalingriemanniandiffusion}, but generally assume that the data already lie on the manifold. \method instead learns the embeddings and the velocity on $\mathbb{S}^{d-1}$ jointly and injects noise via rotations.
\paragraph{Flow Language Models (FLMs).}
A recent line of work treats language modeling as flow matching on continuous representations. Several of these works \citep{sahoo2025diffusionduality, lee2026onesteplanguagemodelingcontinuous, roos2026categoricalflowmaps, potaptchik2026discreteflowmaps} add Gaussian noise to one-hot or simplex representations and decode via $\arg\max$. These approaches materialize dense $L \times |\mathcal{V}|$ arrays at training and sampling time. Fisher-Flow \citep{davis2024fisherflowmatching} maps one-hot vectors to the positive orthant of $\mathbb{S}^{|\mathcal{V}|-1}$ via the Fisher--Rao metric, but this approach does not scale well to language modeling with large vocabularies \citep{jo2025continuousdiffusionmodellanguage}. \method operates on $d$-dimensional token embeddings on $\mathbb{S}^{d-1}$, learned end-to-end. It thus avoids the $|\mathcal{V}|$-dimensional bottleneck and, with the standard DiT backbone, trains faster (\tab{tab:training-cost}).
\paragraph{Representation Learning on the Hypersphere.}Hyperspherical representations are common in contrastive learning, where uniform spread on $\mathbb{S}^{d-1}$ correlates with strong downstream performance \citep{wang2020understandingcontrastiverepresentationlearning}. The cosine distance also outperforms the Euclidean distance for comparing word embeddings \citep{mikolov2013efficientestimationwordrepresentations, pennington2014glove} and for retrieval \citep{reimers2019sentencebertsentenceembeddings, karpukhin2020densepassageretrieval}. A latent prior on $\mathbb{S}^{d-1}$ also stabilizes Variational Autoencoders \citep{davidson2018hypersphericalvariationalautoencoders, xu2018sphericallatentspacesvae}, and normalizing activations and weights to $\mathbb{S}^{d-1}$ improves the stability of AR models \citep{loshchilov2024ngptnormalizedtransformerrepresentation}.
\supp{sec:appendix:related-work} gives a fuller discussion.
\section{Conclusion}
We introduced \method, a Riemannian flow on $\mathbb{S}^{d-1}$ that learns the velocity and the token embeddings jointly. \method is a scalable continuous-flow alternative to one-hot FLMs and is competitive with prior continuous FLMs on OpenWebText. Beyond the formalism, our key contributions include the \arch backbone with normalized activations and the truncated, adaptive noise schedule together with its analysis. On Sudoku, \method performs similarly to prior FLMs. On GSM8K, where other FLMs fail, \method with top-$1$ decoding closes the gap to MDLM and Duo at $T = 1$, although a gap remains at low-temperature sampling ($T = 0.1$). On OpenWebText, \method follows the Gen.\ PPL / Entropy frontier of prior FLMs at high NFE and beats them at low NFE where prior continuous models have an unstable frontier. \method avoids materializing $|\mathcal{V}|$-dimensional one-hot vectors, so with the standard DiT backbone it trains faster than prior FLMs (\tab{tab:training-cost}) and may be easier to scale (though this is speculation and left for future work). Our current implementation of the \arch is slower than the FLM baseline (\sec{sec:limitations}). Our analysis of the sampling dynamics provides a principled heuristic for truncating the noise schedule that works well in practice. At the same time, there remains a clear gap between diffusion models and AR models on GSM8K.
\newpage
\section{Limitations}
\label{sec:limitations}
Continuous diffusion language models, including \method, underperform their discrete counterparts. On GSM8K, \method narrows the gap to discrete diffusion relative to prior FLMs but does not close it at low temperature, and a substantial gap to autoregressive decoding remains (\sec{sec:experiments:tinygsm}). We see two likely causes. First, continuous flows fit a continuous density to discrete data. Prior continuous models commonly close part of this mismatch with self-conditioning \citep{chen2022analogbits, strudel2022selfconditionedembeddingdiffusion}, which we deliberately omit to keep the model simple. Second, the Python solutions in TinyGSM have a strongly left-to-right structure, which favors autoregressive and block-wise diffusion models \citep{arriola2025blockdiffusioninterpolatingautoregressive, deschenaux2026blockgen} over full-sequence diffusion. The truncation threshold $\alpha^\star(\delta)$ is derived from a simplified model (\supp{sec:formal-random-codebook-alpha-crit}) that assumes that the embeddings are distributed uniformly at random on the sphere. The learned embeddings are likely more structured. Therefore, $\alpha^\star(\delta)$ should be treated only as a principled heuristic for the truncation hyperparameter. A more sophisticated model of the sampling dynamics might improve the bound. The \arch follows the design of nGPT for simplicity and trains slower than both the standard DiT and the FLM baseline (\tab{tab:training-cost}), so the training speedup of \method over one-hot FLMs currently holds only with the standard DiT backbone. Improving the throughput of the \arch is an important future direction. The training dynamics of \method should also be studied further, and insights from contrastive representation learning may help. We train a single model per configuration. Training once on TinyGSM costs more than \$300, so we could not afford to train several copies with the same hyperparameters. For the same reason, our experiments are limited to the standard academic scale of the diffusion language modeling literature (170M parameters), and larger-scale experiments are needed to establish whether our conclusions hold as model size grows.
\section{Impact Statement}
\label{sec:impact-statement}
This work advances research on continuous diffusion language models. Like any language modeling research, it carries the standard risk of misuse to generate misinformation or harmful content. Our models are trained at small scale on Sudoku, TinyGSM (synthetic math word problems), and OpenWebText, and remain far below the capabilities of state-of-the-art autoregressive language models. Anyone seeking to cause harm has stronger publicly available models at their disposal. Our contribution is methodological.

\section*{Acknowledgments}
This work has received funding from the Swiss State Secretariat for Education, Research and Innovation (SERI). We acknowledge the SCITAS team at EPFL for providing access to their cluster, and the Swiss National Supercomputing Centre for the Alps platform. We are grateful to Karin G\'etaz for her administrative assistance. We also thank Modal for awarding us a \$10{,}000 Academic Compute Grant (\url{https://modal.com/academics}), which supported the experiments in this work.

\medskip

\bibliographystyle{tmlr}
\bibliography{references}

\newpage
\setcounter{tocdepth}{2}
\tableofcontents
\newpage

\appendix

\section{Background on Flow Matching}
\label{supp:background-fm}
\subsection{The Continuity Equation}
The velocity field $u_t$ and the density $p_t$ are related locally through the \emph{continuity equation}:
\begin{equation}
    \label{eq:continuity}
    \partial_t p_t + \nabla \cdot (p_t\, u_t) = 0.
\end{equation}
This guarantees that matching $u_t$ suffices to match $p_t$, without computing $\phi_t$ or its Jacobian. The Flow Matching (FM) objective regresses $u^\theta_t$ against a target velocity field:
\begin{equation}
    \label{eq:fm-loss}
    \mathcal{L}_{\text{FM}}(\theta) = \mathbb{E}_{t \sim \mathcal{U}[0,1],\, x \sim p_t} \left\| u^\theta_t(x) - u_t(x) \right\|^2.
\end{equation}
In general, $u_t$ and $p_t$ are intractable. CFM replaces them with tractable conditional quantities. Differentiating the conditional flow $x_t = \alpha_t x_1 + (1-\alpha_t) x_0$ yields the conditional velocity field $u_{t|1}(x_t \mid x_1) = \dot{\alpha}_t(x_1 - x_0)$. The marginal velocity field is recovered by averaging under the posterior:
\begin{equation}
    \label{eq:marginal-vf-appendix}
    u_t(x) = \int u_{t|1}(x \mid x_1)\,\frac{p_{t|1}(x \mid x_1)\, q(x_1)}{p_t(x)}\, dx_1 = \int u_{t|1}(x \mid x_1)\, p_{1|t}(x_1 \mid x)\, dx_1.
\end{equation}
The CFM objective has the same gradients and minimizer as $\mathcal{L}_{\text{FM}}$ but is tractable \citep{lipman2023flowmatchinggenerativemodeling, tong2024improvinggeneralizingflowbasedgenerative}.
\subsection{Deriving the Conditional Velocity Fields}
\label{sec:vf-derivation}
Training both Euclidean and Riemannian flow models requires computing the conditional velocity field $u_{t|1}(\mathbf{z}_t \mid \mathbf{z}_1)$ that serves as the regression target in the CFM loss. In both cases, the derivation follows the same recipe: (1) define a conditional flow $\psi_{t|1}$, (2) differentiate with respect to $t$ to obtain $\frac{d\mathbf{z}_t}{dt}$, and (3) eliminate $\mathbf{z}_0$ so that the velocity field is expressed as a function of the current position $\mathbf{z}_t$. During training, any equivalent form of the velocity field can be used as the regression target.
\paragraph{Euclidean Case.}
We define the conditional flow \Eqn{eq:conditional-flow} as the linear interpolation $x_t = \alpha_t\, x_1 + (1 - \alpha_t)\, x_0$. Differentiating with respect to $t$:
\begin{equation}
    \frac{d x_t}{d t} = \dot{\alpha}_t\,(x_1 - x_0).
\end{equation}
By simple algebra, we find $x_0 = (x_t - \alpha_t\, x_1) / (1 - \alpha_t)$. Substituting:
\begin{align}
    u_{t|1}(x_t \mid x_1) &= \dot{\alpha}_t\!\left(x_1 - \frac{x_t - \alpha_t\, x_1}{1 - \alpha_t}\right) = \frac{\dot{\alpha}_t}{1 - \alpha_t}\!\left(x_1(1 - \alpha_t) - x_t + \alpha_t\, x_1\right) \nonumber \\
    &= \boxed{\frac{\dot{\alpha}_t}{1 - \alpha_t}\,(x_1 - x_t).}
\end{align}
For the common choice $\alpha_t = t$, this simplifies to $u_{t|1}(x_t \mid x_1) = (x_1 - x_t) / (1 - t)$. During training, the equivalent form $u_{t|1}(x_t \mid x_1) = \dot{\alpha}_t\,(x_1 - x_0)$ can also be used.
\paragraph{Spherical Case.}
Let $\omega = d_{\mathbb{S}}(\mathbf{z}_0, \mathbf{z}_1) \in (0, \pi)$ be the geodesic distance between the endpoints. The conditional flow \Eqn{eq:spherical-conditional-flow}, written using the SLERP formula \Eqn{eq:slerp}, is
\begin{equation}
    \label{eq:app-xt-slerp}
    \mathbf{z}_t = \frac{\sin((1-\alpha_t)\,\omega)}{\sin\omega}\,\mathbf{z}_0 + \frac{\sin(\alpha_t\,\omega)}{\sin\omega}\,\mathbf{z}_1.
\end{equation}
Differentiating with respect to $t$:
\begin{equation}
    \label{eq:app-dxdt}
    \frac{d\mathbf{z}_t}{dt} = \frac{\dot{\alpha}_t\,\omega}{\sin\omega}\left[-\cos((1-\alpha_t)\,\omega)\,\mathbf{z}_0 + \cos(\alpha_t\,\omega)\,\mathbf{z}_1\right].
\end{equation}
Rewriting \Eqn{eq:app-xt-slerp} gives $\mathbf{z}_0 = \frac{\sin\omega}{\sin((1-\alpha_t)\,\omega)}\,\mathbf{z}_t - \frac{\sin(\alpha_t\,\omega)}{\sin((1-\alpha_t)\,\omega)}\,\mathbf{z}_1$. Substituting into \Eqn{eq:app-dxdt}:
\begin{align}
    \frac{d\mathbf{z}_t}{dt}
    &= \frac{\dot{\alpha}_t\,\omega}{\sin\omega}\left[-\frac{\cos((1{-}\alpha_t)\omega)\,\sin\omega}{\sin((1{-}\alpha_t)\omega)}\,\mathbf{z}_t + \frac{\cos((1{-}\alpha_t)\omega)\,\sin(\alpha_t\omega) + \cos(\alpha_t\omega)\,\sin((1{-}\alpha_t)\omega)}{\sin((1{-}\alpha_t)\omega)}\,\mathbf{z}_1\right].
\end{align}
The numerator of the $\mathbf{z}_1$ coefficient simplifies by the following identity: $\cos((1{-}\alpha_t)\omega)\sin(\alpha_t\omega) + \cos(\alpha_t\omega)\sin((1{-}\alpha_t)\omega) = \sin\omega$. Therefore:
\begin{equation}
    \label{eq:app-dxdt-simplified}
    \frac{d\mathbf{z}_t}{dt} = \frac{\dot{\alpha}_t\,\omega}{\sin((1-\alpha_t)\,\omega)}\left(\mathbf{z}_1 - \cos((1-\alpha_t)\,\omega)\,\mathbf{z}_t\right).
\end{equation}
Recalling the definition of the logarithmic map \Eqn{eq:log-map} with $d_{\mathbb{S}}(\mathbf{z}_t, \mathbf{z}_1) = (1-\alpha_t)\,\omega$, we recognize that \Eqn{eq:app-dxdt-simplified} is proportional to $\log_{\mathbf{z}_t}(\mathbf{z}_1) = \frac{(1-\alpha_t)\,\omega}{\sin((1-\alpha_t)\,\omega)}(\mathbf{z}_1 - \cos((1-\alpha_t)\,\omega)\,\mathbf{z}_t)$. Their ratio is $\dot{\alpha}_t\omega / ((1-\alpha_t)\omega) = \dot{\alpha}_t/(1-\alpha_t)$, giving
\begin{equation}
    \boxed{u_{t|1}(\mathbf{z}_t \mid \mathbf{z}_1) = \frac{\dot{\alpha}_t}{1 - \alpha_t}\,\log_{\mathbf{z}_t}(\mathbf{z}_1).}
\end{equation}
During training, the equivalent form from \Eqn{eq:app-dxdt} can also be used.

\subsection{Continuity Equation for the Hyperspherical Marginal Velocity}
\label{sec:hyperspherical-continuity-equation}
\begin{proposition}[Posterior averaging satisfies the continuity equation]
\label{prop:hyperspherical-continuity}
Let $\mathbf{Z}_1\sim p_1$ and $\mathbf{Z}_0\sim p_0$ be independent points on
$\mathbb{S}^{d-1}$, and let
$\mathbf{Z}_t=\mathrm{SLERP}(\mathbf{Z}_0,\mathbf{Z}_1,\alpha_t)$.
Denote its conditional velocity by
$u_{t|1}(\mathbf{z}\mid\mathbf{z}_1)$, its conditional density by
$p_{t|1}(\mathbf{z}\mid\mathbf{z}_1)$, and its marginal density by $p_t$.
Then the posterior-averaged field
\begin{equation}
    u_t(\mathbf{z})
    = \mathbb{E}\!\left[
        u_{t|1}(\mathbf{Z}_t\mid\mathbf{Z}_1)
        \mid \mathbf{Z}_t=\mathbf{z}
      \right]
    = \int u_{t|1}(\mathbf{z}\mid\mathbf{z}_1)
      p_{1|t}(\mathbf{z}_1\mid\mathbf{z})\,d\mathbf{z}_1
    \label{eq:app-hyperspherical-marginal-vf}
\end{equation}
satisfies the continuity equation for $p_t$:
\begin{equation}
    \partial_t p_t
    + \operatorname{div}_{\!\mathbb{S}}(p_t u_t)=0.
    \label{eq:app-hyperspherical-continuity-density}
\end{equation}
\end{proposition}

\begin{proof}
For a fixed endpoint $\mathbf{z}_1$, the conditional SLERP transports $p_0$
with velocity $u_{t|1}(\cdot\mid\mathbf{z}_1)$. Conservation of probability
along this conditional flow gives
\begin{equation}
    \partial_t p_{t|1}(\mathbf{z}\mid\mathbf{z}_1)
    + \operatorname{div}_{\!\mathbb{S}}\!\left(
        p_{t|1}(\mathbf{z}\mid\mathbf{z}_1)
        u_{t|1}(\mathbf{z}\mid\mathbf{z}_1)
      \right)=0.
    \label{eq:app-hyperspherical-conditional-continuity}
\end{equation}
The marginal density is the mixture of these conditional densities,
\begin{equation}
    p_t(\mathbf{z})
    = \int p_{t|1}(\mathbf{z}\mid\mathbf{z}_1)
      p_1(\mathbf{z}_1)\,d\mathbf{z}_1.
    \label{eq:app-hyperspherical-marginal-density}
\end{equation}
Moreover, Bayes' rule and \Eqn{eq:app-hyperspherical-marginal-vf} imply
\begin{equation}
    p_t(\mathbf{z})u_t(\mathbf{z})
    = \int p_{t|1}(\mathbf{z}\mid\mathbf{z}_1)
      p_1(\mathbf{z}_1)
      u_{t|1}(\mathbf{z}\mid\mathbf{z}_1)\,d\mathbf{z}_1.
    \label{eq:app-hyperspherical-marginal-flux}
\end{equation}
Differentiating \Eqn{eq:app-hyperspherical-marginal-density}, taking the
divergence of \Eqn{eq:app-hyperspherical-marginal-flux}, and using linearity
therefore gives
\begin{align}
    \partial_t p_t(\mathbf{z})
    + \operatorname{div}_{\!\mathbb{S}}(p_tu_t)(\mathbf{z})
    &= \int p_1(\mathbf{z}_1)\left[
      \partial_t p_{t|1}(\mathbf{z}\mid\mathbf{z}_1)
      + \operatorname{div}_{\!\mathbb{S}}\!\left(
        p_{t|1}(\mathbf{z}\mid\mathbf{z}_1)
        u_{t|1}(\mathbf{z}\mid\mathbf{z}_1)
      \right)\right]d\mathbf{z}_1 \\
    &= 0,
\end{align}
where the final equality follows from
\Eqn{eq:app-hyperspherical-conditional-continuity}.
\end{proof}

For a single S-FLM token, $\mathbf{z}_1=\hat{\mathbf{e}}_{\x}$ and
\Eqn{eq:spherical-conditional-vf} gives
\begin{equation}
    u_t(\mathbf{z})
    = \frac{\dot{\alpha}_t}{1-\alpha_t}
      \sum_{v\in\mathcal{V}}
      p_{1|t}(\x=v\mid\mathbf{z})
      \log_{\mathbf{z}}(\hat{\mathbf{e}}_v).
    \label{eq:app-sflm-exact-velocity}
\end{equation}
At a population optimum $\theta^\star$, the Cross-Entropy recovers the true posterior,
$p^{\theta^\star}_{1|t}(v\mid\mathbf{z})=p_{1|t}(\x=v\mid\mathbf{z})$. Thus,
\Eqn{eq:app-sflm-exact-velocity} is the posterior-averaged velocity from
Proposition~\ref{prop:hyperspherical-continuity} and satisfies the continuity
equation.

\section{Additional Details}

\subsection{Training and Sampling Pseudocode}
\label{sec:pseudocode}
\algo{alg:training} shows the training pseudocode for \method, \algo{alg:sampling} the deterministic sampler, and \algo{alg:sampling-stochastic} the stochastic sampler that replaces the posterior-weighted velocity by the log map toward a single token sampled from $p^\theta_{1|t}$.

\begin{algorithm}[H]
\caption{Training}
\label{alg:training}
\begin{algorithmic}[1]
\REQUIRE Dataset of token sequences, embedding table $\mathbf{E}$, scheduling function $\alpha_t$
\REPEAT
    \STATE Sample sequence $\x \sim p_1$, time $t \sim \mathcal{U}[0,1]$
    \FOR{$\ell = 1$ \TO $L$}
        \STATE $\mathbf{z}_1^\supl \leftarrow \mathbf{E}[\x^\supl] / \|\mathbf{E}[\x^\supl]\|$ \Comment{Embed and normalize}
        \STATE $\boldsymbol{\epsilon}^\supl \sim \mathcal{N}(\mathbf{0}, \mathbf{I}_d);\quad \mathbf{z}_0^\supl \leftarrow \boldsymbol{\epsilon}^\supl / \|\boldsymbol{\epsilon}^\supl\|$ \Comment{Sample noise on $\mathbb{S}^{d-1}$}
        \STATE $\mathbf{z}_t^\supl \leftarrow \mathrm{SLERP}(\mathbf{z}_0^\supl, \mathbf{z}_1^\supl, \alpha_t)$ \Comment{Construct noisy latent}
    \ENDFOR
    \STATE Compute $\mathcal{L}_{\text{CE}} = -\sum_\ell \log p^\theta_{1|t}(\x^\supl \mid \mathbf{z}_t)$ \Comment{Cross-Entropy loss}
    \STATE Update $\theta$ and $\mathbf{E}$ by gradient descent on $\mathcal{L}_{\text{CE}}$
\UNTIL{converged}
\end{algorithmic}
\end{algorithm}
\begin{algorithm}[H]
\caption{Sampling}
\label{alg:sampling}
\begin{algorithmic}[1]
\REQUIRE Denoiser $p^\theta_{1|t}$, embedding table $\mathbf{E}$ with rows $\mathbf{e}_v$ and unit-norm versions $\hat{\mathbf{e}}_v = \mathbf{e}_v / \|\mathbf{e}_v\|$, number of steps $N$, decoder $p_{\text{dec}}$
\FOR{$\ell = 1$ \TO $L$}
    \STATE $\boldsymbol{\epsilon}^\supl \sim \mathcal{N}(\mathbf{0}, \mathbf{I}_d);\quad \mathbf{z}_0^\supl \leftarrow \boldsymbol{\epsilon}^\supl / \|\boldsymbol{\epsilon}^\supl\|$ \Comment{Sample noise on $\mathbb{S}^{d-1}$}
\ENDFOR
\FOR{$n = 0$ \TO $N{-}1$}
    \STATE $s_n \leftarrow (\alpha_{t_{n+1}} - \alpha_{t_n})\,/\,(1 - \alpha_{t_n})$ \Comment{Step size}
    \STATE Compute $p^\theta_{1|t_n}(\cdot \mid \mathbf{z}_{t_n})$ \Comment{One forward pass}
    \FOR{$\ell = 1$ \TO $L$}
        \STATE $\bar{\mathbf{v}}^\supl \leftarrow \sum_{v \in \mathcal{V}} p^\theta_{1|t_n}(v \mid \mathbf{z}_{t_n}) \, \log_{\mathbf{z}_{t_n}^\supl}\!\left(\hat{\mathbf{e}}_v\right)$ \Comment{Posterior-weighted velocity}
        \STATE $\mathbf{z}_{t_{n+1}}^\supl \leftarrow \exp_{\mathbf{z}_{t_n}^\supl}\!\left(s_n \cdot \bar{\mathbf{v}}^\supl\right)$ \Comment{Geodesic Euler step}
    \ENDFOR
\ENDFOR
\STATE $\x \sim p_{\text{dec}}(\cdot \mid \mathbf{z}_1)$ \Comment{Decode (argmax or AR)}
\RETURN $\x$
\end{algorithmic}
\end{algorithm}
\begin{algorithm}[H]
\caption{Stochastic sampling}
\label{alg:sampling-stochastic}
\begin{algorithmic}[1]
\REQUIRE Denoiser $p^\theta_{1|t}$, embedding table $\mathbf{E}$ with rows $\mathbf{e}_v$ and unit-norm versions $\hat{\mathbf{e}}_v = \mathbf{e}_v / \|\mathbf{e}_v\|$, number of steps $N$, decoder $p_{\text{dec}}$
\FOR{$\ell = 1$ \TO $L$}
    \STATE $\boldsymbol{\epsilon}^\supl \sim \mathcal{N}(\mathbf{0}, \mathbf{I}_d);\quad \mathbf{z}_0^\supl \leftarrow \boldsymbol{\epsilon}^\supl / \|\boldsymbol{\epsilon}^\supl\|$ \Comment{Sample noise on $\mathbb{S}^{d-1}$}
\ENDFOR
\FOR{$n = 0$ \TO $N{-}1$}
    \STATE $s_n \leftarrow (\alpha_{t_{n+1}} - \alpha_{t_n})\,/\,(1 - \alpha_{t_n})$ \Comment{Step size}
    \STATE Compute $p^\theta_{1|t_n}(\cdot \mid \mathbf{z}_{t_n})$ \Comment{One forward pass}
    \FOR{$\ell = 1$ \TO $L$}
        \STATE Sample $\hat{\x}^\supl \sim p^\theta_{1|t_n}(\cdot \mid \mathbf{z}_{t_n})$ \Comment{Posterior sample}
        \STATE $\bar{\mathbf{v}}^\supl \leftarrow \log_{\mathbf{z}_{t_n}^\supl}\!\left(\hat{\mathbf{e}}_{\hat{\x}^\supl}\right)$ \Comment{Velocity toward sampled token}
        \STATE $\mathbf{z}_{t_{n+1}}^\supl \leftarrow \exp_{\mathbf{z}_{t_n}^\supl}\!\left(s_n \cdot \bar{\mathbf{v}}^\supl\right)$ \Comment{Geodesic Euler step}
    \ENDFOR
\ENDFOR
\STATE $\x \sim p_{\text{dec}}(\cdot \mid \mathbf{z}_1)$ \Comment{Decode (argmax or AR)}
\RETURN $\x$
\end{algorithmic}
\end{algorithm}

\subsection{Adaptive Noise Schedule}
\label{sec:adaptive-schedule}
\paragraph{Motivation.}
The denoising task is not as meaningful at every noise level. At low noise levels, the task is trivial since the noisy embeddings are very close to the clean embeddings \Eqn{eq:alpha-crit}. At high noise levels, the latent $\mathbf{z}_t^\supl$ is close to pure noise hence carries little signal about the clean token. In both cases the denoiser $p^\theta_{1|t}$ has little to learn. The informative region lies in between, where the model can extract signal from noise and the loss drops fastest as a function of $t$. Inspired by InfoNoise \citep{raya2026informationguidednoiseallocation}, we focus training on regions where the loss derivative $|d\mathcal{L}/dt|$ is largest.
\begin{algorithm}[H]
\caption{Adaptive noise schedule refit}
\label{alg:adaptive-schedule}
\begin{algorithmic}[1]
\REQUIRE buffer $\{(t_i, \mathcal{L}_i)\}$, base schedule $\alpha^{\text{base}}$, grid $\{t_j\}_{j=1}^N \subset [0,1]$, EMA rate $\beta$, uniform mix $\mu$, ridge $\lambda$, refit count $n$
\STATE Fit a spline $\hat{\mathcal{L}}(t)$ to $(t_i, \mathcal{L}_i)$ via ridge regression ($\lambda$)
\STATE $g(t_j) \leftarrow \max\{d\hat{\mathcal{L}}/dt\,(t_j),\, 0\}$ \Comment{Clamp: loss should grow with noise}
\STATE $w(t_j) \leftarrow (1 - \mu)\, g(t_j) + \mu$ \Comment{Ensure the CDF is invertible}
\STATE $F(t_j) \leftarrow \sum_{j' \le j} w(t_{j'}) \big/ \sum_{j'} w(t_{j'})$ \Comment{Empirical CDF (discrete points)}
\STATE $\tilde{t}_j \leftarrow F^{-1}(t_j)$ via PCHIP interpolation \Comment{Invert CDF}
\STATE $\bar{\alpha}_j \leftarrow \beta\, \bar{\alpha}_j + (1-\beta)\, \alpha^{\text{base}}(\tilde{t}_j)$ \Comment{EMA update}
\STATE $\alpha^{\text{final}}_j \leftarrow \bar{\alpha}_j / (1 - \beta^n)$ \Comment{Bias correction}
\STATE Store $\alpha^{\text{adapt}}_t$ as a PCHIP spline through $\{(t_j, \alpha^{\text{final}}_j)\}$
\end{algorithmic}
\end{algorithm}
During training we append $(t, \mathcal{L})$ pairs to the buffer at every step and invoke \algo{alg:adaptive-schedule} every $R$ steps after a warmup of $W = 1000$ steps. The EMA is essential to reduce oscillations in the schedule, and the Adam-style bias correction $1/(1-\beta^n)$ \citep{kingma2017adammethodstochasticoptimization} prevents the early-training estimate from being biased toward zero. By default we use $R = 50$, buffer size $R \cdot \text{batch size}$, $\beta = 0.9$, and $\mu = 10^{-3}$. All operations run on CPU in \texttt{numpy}, so the impact on training throughput is minimal. Empirically, the noise schedule stabilizes quickly.
\paragraph{Implementation.}
We fit the loss profile using a spline (\texttt{sklearn.preprocessing.SplineTransformer}) with ridge regression (\texttt{sklearn.linear\_model.Ridge}). For interpolating the inverse CDF $F^{-1}$ and the final noise schedule $\alpha^{\text{adapt}}_t$, we use \texttt{scipy.interpolate.PchipInterpolator}, which builds a \emph{Piecewise Cubic Hermite Interpolating Polynomial}~\citep{fritsch1984method} that preserves the monotonicity of the data.

\paragraph{Relation to InfoNoise.}
Our adaptive schedule is inspired by InfoNoise \citep{raya2026informationguidednoiseallocation}, which also adapts the noise schedule online from the value of the loss. We differ in several ways.
\begin{itemize}
  \item \textbf{Importance signal.} InfoNoise estimates the conditional entropy rate $\dot{H}[\mathbf{x}_0 \mid \mathbf{x}_\sigma] \propto \mathrm{mmse}(\sigma)/\sigma^3$ via the I--MMSE identity. We use $|d\hat{\mathcal{L}}/dt|$ directly on the Cross-Entropy profile.
  \item \textbf{Estimator.} InfoNoise splits the $\sigma$-range into bins, each with a FIFO buffer for recent losses and an EMA on the MSE estimate. We fit a spline to $(t, \mathcal{L})$ pairs via ridge regression. This removes the need to choose a number of bins.
  \item \textbf{Stabilization.} InfoNoise uses pivot calibration and gating to suppress boundary artifacts. We enforce a minimum CDF increment $\mu$ to keep the CDF invertible and stabilize the schedule with an EMA, in addition to truncating the noise schedule range \Eqn{eq:alpha-crit}.
  \item \textbf{Loss weight.} InfoNoise separates sampling density $\pi(\sigma)$ from loss weight $w(\sigma)$ via the effective weight $\phi = \pi \cdot w$. We use unweighted Cross-Entropy ($w \equiv 1$), so the sampling density matches effective weight directly.
\end{itemize}
\subsection{Hyperspherical Backbone Architecture}
\label{sec:ngpt-arch}
In this section, we present the hyperspherical denoising backbone in more details. The architecture is adapted from nGPT \citep{loshchilov2024ngptnormalizedtransformerrepresentation} for \method. The main differences are that (1) we use bidirectional attention, (2) we use GELU activations in the MLP, since the other AR / diffusion denoisers also use GELU, (3) we make the residual gates time-conditional. As in nGPT, every weight matrix has unit-norm vectors along the embedding-dimension axis. We enforce this at initialization and optionally re-apply the projection after every optimizer step.
\paragraph{Notation.}
Let $\mathrm{Norm}(\mathbf{u}) = \mathbf{u}/\max(\|\mathbf{u}\|_2, \varepsilon)$ with $\varepsilon = 10^{-6}$ project a vector onto the unit sphere along its last axis. We write $d$ for the embedding dimension, $H$ for the number of attention heads, $d_k = d/H$, and $N$ for the number of transformer blocks. The base scale $b = 1/\sqrt{d}$ appears in several rescaling parameters. All per-dimension scales ($\mathbf{s}_{qk}, \mathbf{s}_{fc}, \mathbf{s}_z$) and residual gates ($\boldsymbol{\gamma}_A, \boldsymbol{\gamma}_M$) introduced below are learnable.
\paragraph{Normalized Attention.}
The multi-head attention uses bias-free linear layers: $\mathbf{Q} = \mathbf{h}\mathbf{W}_Q$, $\mathbf{K} = \mathbf{h}\mathbf{W}_K$, $\mathbf{V} = \mathbf{h}\mathbf{W}_V$, split into $H$ heads. RoPE \citep{su2023roformerenhancedtransformerrotary} is applied to $\mathbf{Q}, \mathbf{K}$. We then apply the normalization and rescaling:
\begin{equation*}
    \tilde{\mathbf{s}}_{qk} = \mathbf{s}_{qk} \cdot s_{qk}^{\mathrm{init}}/s_{qk}^{\mathrm{scale}},
    \qquad
    \mathbf{Q}' = \tilde{\mathbf{s}}_{qk} \odot \mathrm{Norm}(\mathbf{Q}),
    \quad
    \mathbf{K}' = \tilde{\mathbf{s}}_{qk} \odot \mathrm{Norm}(\mathbf{K}),
\end{equation*}
with $s_{qk}^{\mathrm{init}} = 1$ and $s_{qk}^{\mathrm{scale}} = b$. The softmax scaling follows from the variance of the inner product $\langle \hat{\mathbf{q}}, \hat{\mathbf{k}} \rangle$ between two unit-norm vectors $\hat{\mathbf{q}} = \mathrm{Norm}(\mathbf{Q})$ and $\hat{\mathbf{k}} = \mathrm{Norm}(\mathbf{K})$. Assume the components of $\mathbf{Q}, \mathbf{K}$ are i.i.d.\ zero-mean Gaussian. After projection by $\mathrm{Norm}$, $\hat{\mathbf{q}}$ and $\hat{\mathbf{k}}$ are independent samples from $\mathrm{Unif}(\mathbb{S}^{d_k - 1})$. By rotational symmetry and $\|\mathbf{u}\|^2 = 1$, any $\mathbf{u} \sim \mathrm{Unif}(\mathbb{S}^{d_k - 1})$ satisfies $\mathbb{E}[u_i] = 0$ and $\mathbb{E}[u_i u_j] = \delta_{ij}/d_k$, hence $\mathbb{E}[\langle \hat{\mathbf{q}}, \hat{\mathbf{k}} \rangle] = 0$ and $\mathrm{Var}[\langle \hat{\mathbf{q}}, \hat{\mathbf{k}} \rangle] = 1/d_k$. Following nGPT, we therefore set the softmax scale to $\sqrt{d_k}$:
\begin{equation*}
    \mathbf{O} = \mathrm{softmax}\!\bigl(\sqrt{d_k}\, \mathbf{Q}' \mathbf{K}'^\top\bigr) \mathbf{V},
    \qquad
    \mathbf{h}_A = \mathbf{O}\,\mathbf{W}_O.
\end{equation*}
\paragraph{Normalized MLP.}
The MLP is similar to the MLP in discrete-diffusion DiTs \citep{peebles2023scalablediffusionmodelstransformers, lou2024discretediffusionmodelingestimating}, with the addition of a per-dimension scale $\mathbf{s}_{fc}$ on the hidden activations:
\begin{equation*}
    \mathbf{h}_M = \mathrm{GELU}\bigl(\sqrt{d}\, \mathbf{s}_{fc} \odot (\mathbf{h}\,\mathbf{W}_{fc})\bigr)\,\mathbf{W}_O^M.
\end{equation*}
As in the previous section, the $\sqrt{d}$ factor restores unit variance to the inputs of GELU.
\paragraph{Timestep Injection.}
Let $\boldsymbol{\tau}(t) = \mathrm{SiLU}(\mathrm{TimestepEmbedder}(\sigma(t))) \in \mathbb{R}^{d_{\mathrm{cond}}}$ be the timestep embedding (sinusoidal embedding followed by a two-layer MLP). Each transformer block has its own zero-initialized linear map $\mathbf{W}_{\boldsymbol{\delta}} \in \mathbb{R}^{2d \times d_{\mathrm{cond}}}$ that computes per-dimension biases
$(\boldsymbol{\delta}_A(t), \boldsymbol{\delta}_M(t)) = \mathbf{W}_{\boldsymbol{\delta}} \boldsymbol{\tau}(t)$,
for the residual update (next paragraph).
\paragraph{Residual Update.}
The block updates the hidden state by interpolating with the (normalized) attention output, and then with the (normalized) MLP output. Each interpolation is gated by learnable per-dimension parameters $\boldsymbol{\gamma}_A, \boldsymbol{\gamma}_M \in \mathbb{R}^d$, and renormalized:
\begin{align*}
    \mathbf{h} &\leftarrow \mathrm{Norm}\Bigl(\mathrm{Norm}(\mathbf{h}) + \tilde{\boldsymbol{\gamma}}_A \odot \bigl(\mathrm{Norm}(\mathbf{h}_A) - \mathrm{Norm}(\mathbf{h})\bigr)\Bigr), \\
    \mathbf{h} &\leftarrow \mathrm{Norm}\Bigl(\mathrm{Norm}(\mathbf{h}) + \tilde{\boldsymbol{\gamma}}_M \odot \bigl(\mathrm{Norm}(\mathbf{h}_M) - \mathrm{Norm}(\mathbf{h})\bigr)\Bigr).
\end{align*}
The effective gate is $\tilde{\boldsymbol{\gamma}}_X = |\boldsymbol{\gamma}_X \cdot (\gamma^{\mathrm{init}}/\gamma^{\mathrm{scale}}) + \boldsymbol{\delta}_X(t)|$ with $\gamma^{\mathrm{init}} = 0.05$ and $\gamma^{\mathrm{scale}} = b$, for $X \in \{A, M\}$, where $\boldsymbol{\delta}_X(t)$ is the time modulation defined above. Except for the time modulation $\boldsymbol{\delta}_X(t)$, the design and hyperparameters, are carried from nGPT. The absolute value ensures the gate is non-negative, so the update always pulls $\mathbf{h}$ toward the block output.
\paragraph{Output Head.}
We project the output of the last block through a row-normalized linear map $\mathbf{W}_{\mathrm{lm}} \in \mathbb{R}^{|\mathcal{V}| \times d}$ and rescale by a learnable scale $\mathbf{s}_z \in \mathbb{R}^{|\mathcal{V}|}$:
\begin{equation*}
    \tilde{\mathbf{s}}_z = \mathbf{s}_z \cdot s_z^{\mathrm{init}}/s_z^{\mathrm{scale}},
    \qquad
    \boldsymbol{\ell} = \tilde{\mathbf{s}}_z \odot (\mathbf{h}\,\mathbf{W}_{\mathrm{lm}}^\top),
\end{equation*}
with $s_z^{\mathrm{init}} = 1$ and $s_z^{\mathrm{scale}} = b$. The rescaling is necessary because $\mathbf{h}\mathbf{W}_{\mathrm{lm}}^\top$ contains inner products of unit vectors and is therefore bounded in $[-1, 1]^{|\mathcal{V}|}$. Without $\mathbf{s}_z$ the Cross-Entropy would saturate near a uniform distribution.

\subsection{Step Size with the Euler Sampler}
\label{sec:step-size-derivation}
The marginal velocity from \Eqn{eq:flm-marginal-vf} is $u_t^\supl = \frac{\dot{\alpha}_t}{1-\alpha_t}\,\bar{u}_t^\supl$, where $\bar{u}_t^\supl = \sum_v p_{1|t}(\x^\supl = v \mid \mathbf{z}_t)\,\log_{\mathbf{z}_t^\supl}(\hat{\mathbf{e}}_v)$. During sampling, to take an Euler step of size $\Delta t$ from time time $t_n$ to $t_{n+1}$, we compute
\begin{equation}
    \mathbf{z}_{t_{n+1}}^\supl = \exp_{\mathbf{z}_{t_n}^\supl}\!\left(\Delta t \cdot u_{t_n}^\supl\right) = \exp_{\mathbf{z}_{t_n}^\supl}\!\left(\frac{\dot{\alpha}_{t_n}\,\Delta t}{1-\alpha_{t_n}} \cdot \bar{u}_{t_n}^\supl\right).
\end{equation}
Using the first-order approximation $\dot{\alpha}_{t_n}\,\Delta t \approx \alpha_{t_{n+1}} - \alpha_{t_n}$, the step size applied to $\bar{u}_t^\supl$ is
\begin{equation}
    \label{eq:step-size}
    s_n = \frac{\alpha_{t_{n+1}} - \alpha_{t_n}}{1 - \alpha_{t_n}},
\end{equation}
so that $\mathbf{z}_{t_{n+1}}^\supl = \exp_{\mathbf{z}_{t_n}^\supl}(s_n \cdot \bar{u}_{t_n}^\supl)$. For the linear schedule $\alpha_t = t$ with uniform time steps $t_n = n/N$, this gives $s_n = 1/(N-n)$.
\subsection{Re-Normalizing Embeddings}
\label{sec:forced-normalization}
We show that when embeddings are normalized during the forward pass, the norm of the embeddings grow at every optimization step, and thus we obtain a decay of the effective learning rate (since the angular updates become smaller). Our argument is similar to prior work \citep{karras2024analyzingimprovingtrainingdynamics}.

\paragraph{Setup.} Let $\mathbf{e} \in \mathbb{R}^d$ be an \emph{unconstrained} embedding vector, and let $\hat{\mathbf{e}} = \mathbf{e} / \|\mathbf{e}\|$ be the normalized version used in the forward pass. The loss $\mathcal{L}$ depends on $\mathbf{e}$ only through $\hat{\mathbf{e}}$.

\paragraph{Jacobian of the Normalization.} The Jacobian of $\hat{\mathbf{e}}$ with respect to $\mathbf{e}$ is
\begin{equation}
    \frac{\partial \hat{\mathbf{e}}}{\partial \mathbf{e}} = \frac{1}{\|\mathbf{e}\|}\left(\mathbf{I} - \hat{\mathbf{e}}\,\hat{\mathbf{e}}^\top\right).
\end{equation}
By the chain rule, the gradient of $\mathcal{L}$ with respect to the unnormalized embedding is
\begin{equation}
    \label{eq:grad-unnormalized}
    \nabla_\mathbf{e} \mathcal{L} = \left(\frac{\partial \hat{\mathbf{e}}}{\partial \mathbf{e}}\right)^\top \nabla_{\hat{\mathbf{e}}} \mathcal{L} = \frac{1}{\|\mathbf{e}\|}\left(\mathbf{I} - \hat{\mathbf{e}}\,\hat{\mathbf{e}}^\top\right) \nabla_{\hat{\mathbf{e}}} \mathcal{L},
\end{equation}
where we used the fact that $\mathbf{I} - \hat{\mathbf{e}}\,\hat{\mathbf{e}}^\top$ is symmetric.

\paragraph{The Gradient Is Orthogonal to $\mathbf{e}$.} The matrix $\mathbf{P} = \mathbf{I} - \hat{\mathbf{e}}\,\hat{\mathbf{e}}^\top$ projects onto the subspace orthogonal to $\hat{\mathbf{e}}$. Since $\mathbf{e} = \|\mathbf{e}\|\,\hat{\mathbf{e}}$:
\begin{align}
    \mathbf{e}^\top \nabla_\mathbf{e} \mathcal{L}
    &= \frac{1}{\|\mathbf{e}\|}\,\mathbf{e}^\top\!\left(\mathbf{I} - \hat{\mathbf{e}}\,\hat{\mathbf{e}}^\top\right)\nabla_{\hat{\mathbf{e}}} \mathcal{L} \nonumber \\
    &= \frac{1}{\|\mathbf{e}\|}\left(\mathbf{e}^\top - \underbrace{\mathbf{e}^\top\hat{\mathbf{e}}}_{=\,\|\mathbf{e}\|}\,\hat{\mathbf{e}}^\top\right)\nabla_{\hat{\mathbf{e}}} \mathcal{L} \nonumber \\
    &= \frac{1}{\|\mathbf{e}\|}\left(\mathbf{e}^\top - \|\mathbf{e}\|\,\hat{\mathbf{e}}^\top\right)\nabla_{\hat{\mathbf{e}}} \mathcal{L} \nonumber \\
    &= \frac{1}{\|\mathbf{e}\|}\left(\mathbf{e}^\top - \mathbf{e}^\top\right)\nabla_{\hat{\mathbf{e}}} \mathcal{L} = 0,
\end{align}
where we used $\|\mathbf{e}\|\,\hat{\mathbf{e}} = \mathbf{e}$.

\paragraph{Norm Growth.} Since $\nabla_\mathbf{e}\mathcal{L} \perp \mathbf{e}$, a gradient step $\mathbf{e}' = \mathbf{e} - \eta\,\nabla_\mathbf{e}\mathcal{L}$ gives, by the Pythagorean theorem:
\begin{equation}
    \|\mathbf{e}'\|^2 = \|\mathbf{e}\|^2 + \eta^2\,\|\nabla_\mathbf{e}\mathcal{L}\|^2 > \|\mathbf{e}\|^2.
\end{equation}
The norm grows at every step, which induces an effective learning rate decay on the embedding table.

\paragraph{Fix.} After each optimization step, we can re-project $\mathbf{e} \leftarrow \mathbf{e} / \|\mathbf{e}\|$ (not through the computational graph, but by directly modifying the parameters). This ensures that $\|\mathbf{e}\| = 1$ and that there is no implicit learning rate annealing\citep{karras2024analyzingimprovingtrainingdynamics}. Empirically, we find that the learning rate annealing on the embedding table is beneficial.

\subsection{Training Cost}
\label{app:training-cost}
Training on Sudokus is fast and cheap. We use a single L40S GPU and train for less than 2 hours. Using the on-demand price on RunPod (\url{https://www.runpod.io}), training costs less than \$2.

\tab{tab:training-cost} shows the step per second and total cost on TinyGSM and OpenWebText. We train on H100 GPUs with bfloat16 mixed precision. We use 8 H100s for TinyGSM, and 16 H100s on OpenWebtext. We train for 250k steps on TinyGSM (\sec{sec:experiments:tinygsm}) and 1M on OpenWebText (\sec{sec:experiments:owt}).
\begin{table}[t]
\centering
\caption{\textbf{Training cost} on TinyGSM and OpenWebText. The total duration and GPU-hours are derived from the step/sec. We did not investigate deeply why \method trains faster than MDLM. We did not try to hyper-optimize our implementations. For MDLM, CANDI and \method, we import their original implementation, based on the Duo codebase (\url{https://github.com/s-sahoo/duo}), since we also use it. It is plausible that MDLM / Duo can be made to train as fast as \method. Observe that the FLM baseline is slower than \method. This is expected since it materializes dense one-hot-diffused arrays, and multiplies them with the embedding table, which is costly.}
\label{tab:training-cost}
\small
\setlength{\tabcolsep}{4pt}
\renewcommand{\arraystretch}{1.05}
\begin{tabular}{l r r r r r r}
\toprule
 & \multicolumn{2}{c}{Steps/sec} & \multicolumn{2}{c}{Duration (h)} & \multicolumn{2}{c}{GPU-hours} \\
\cmidrule(lr){2-3} \cmidrule(lr){4-5} \cmidrule(lr){6-7}
Model & TinyGSM & OWT & TinyGSM & OWT & TinyGSM & OWT \\
\midrule
AR             & $5.4$ & $6.0$ & $12.8$ & $46.6$ & $102.1$ & $745.7$ \\
MDLM           & $3.0$ & $3.5$ & $22.8$ & $80.3$ & $182.7$ & $1284.5$ \\
Duo            & $3.0$ & $3.5$ & $22.8$ & $80.3$ & $182.7$ & $1284.5$ \\
CANDI          & $3.0$ & $2.5$ & $23.5$ & $108.9$ & $188.3$ & $1742.9$ \\
FLM            & $3.2$ & $3.3$ & $21.7$ & $84.2$ & $173.6$ & $1346.8$ \\
\method                          & $4.2$ & $4.5$ & $16.5$ & $61.7$ & $132.3$ & $987.7$ \\
\method + adapt.                 & $4.2$ & $4.5$ & $16.5$ & $61.7$ & $132.3$ & $987.7$ \\
\method + adapt. + \arch  & $2.6$ & $2.8$ & $26.7$ & $99.2$ & $213.7$ & $1587.3$ \\
\bottomrule
\end{tabular}
\end{table}

\section{Additional Results}
\subsection{Sequence Length Using Different Tokenizers}
\label{sec:appendix:tokenlen}
\begin{figure}[t]
    \centering
    \includegraphics[width=\linewidth]{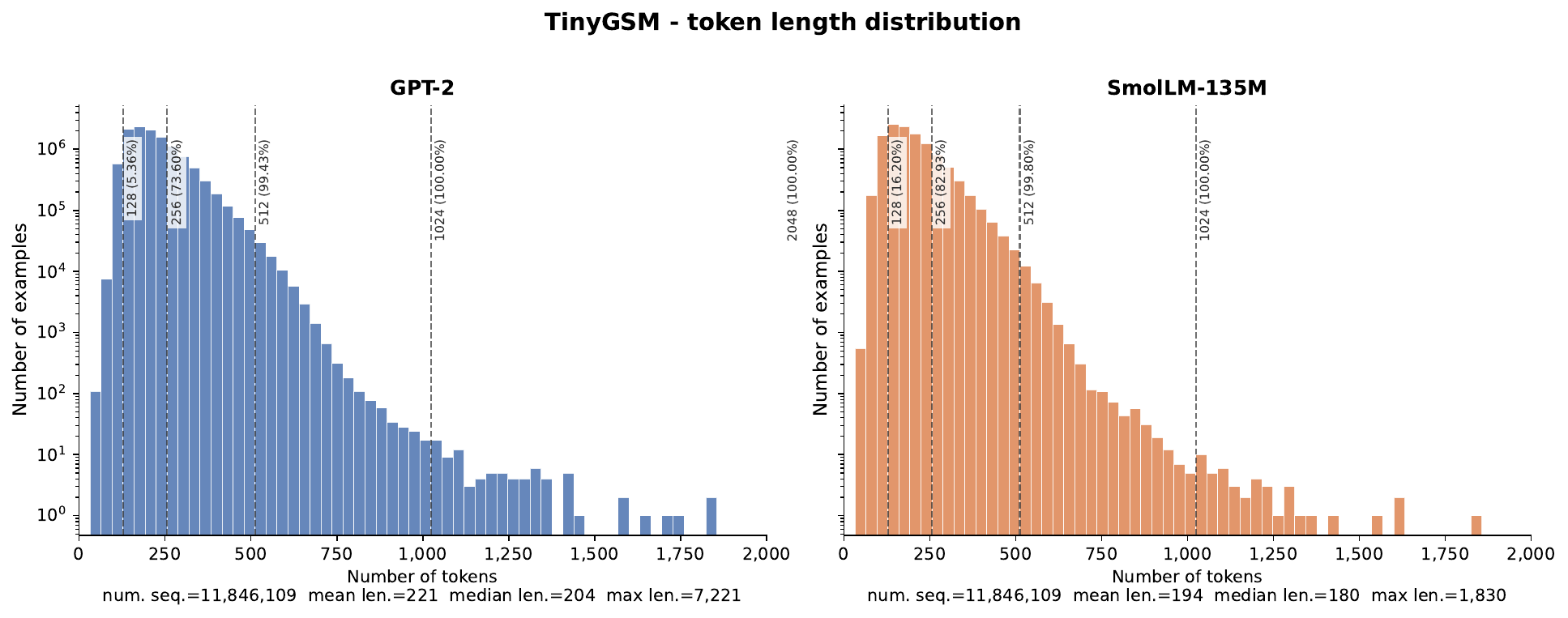}
    \caption{Distribution of tokenized sequence lengths on TinyGSM, under the GPT-2 tokenizer (left) and the \texttt{SmolLM-135M} tokenizer (right). The SmolLM tokenizer yields shorter sequences on average (median 204 vs.\ 180 tokens), which allows us to fit more problems in a fixed context length of 512 tokens.}
    \label{fig:tokenlen-smollm-gpt2}
\end{figure}
In \Fig{fig:tokenlen-smollm-gpt2}, we plot the histogram of length of the tokenized examples in TinyGSM \citep{liu2023tinygsm}, when tokenized with the GPT-2 tokenizer \citep{Radford2019LanguageMA} or the SmolLM tokenizer \citep{allal2025smollm2smolgoesbig}. Since the SmolLM tokenizer was trained on code, unlike GPT-2, we see that it offers better compression. Therefore, for the experiments on TinyGSM, we use the SmolLM tokenizer.

\subsection{Analysis Under the Random Codebook Model}
\label{sec:formal-random-codebook-alpha-crit}
To derive \Eqn{eq:alpha-crit}, we use the standard concentration bound for inner products of uniform random vectors on the sphere \citep[Theorem 3.4.5]{vershynin2018high}. For $\mathbf{X}, \mathbf{Y} \in \mathbb{S}^{d-1}$ with at least one of them sampled from $\mathcal{U}(\mathbb{S}^{d-1})$ and the other either fixed or independently sampled, for every $\epsilon > 0$:
\begin{equation}
    \label{eq:app-one-sided-sphere-tail}
    \mathbb{P}\!\left(\mathbf{X}^\top \mathbf{Y} > \epsilon\right)
    \leq
    2 \exp\!\left(- \frac{d \epsilon^2}{2}\right).
\end{equation}
The probability decreases exponentially in the dimension $d$.

\paragraph{Closed-Form for the Sampling Dynamics.}
Let $\mathbf{z}_\alpha = \mathrm{SLERP}(\mathbf{z}_0, \hat{\mathbf{e}}_j, \alpha)$ for $\alpha \in [0,1]$. Then, by definition of SLERP, we have
\begin{equation}
    \label{eq:app-slerp-clean-score}
    \langle \mathbf{z}_\alpha, \hat{\mathbf{e}}_j \rangle = \cos\!\bigl(\omega(1-\alpha)\bigr),
\end{equation}

where $\omega$ is the angle between $\mathbf{z}_0$ and $\hat{\mathbf{e}}_j$. Indeed, we recall the standard trigonometric identities
\begin{equation}
    \sin(a-b) = \sin(a)\cos(b) - \cos(a)\sin(b)
\end{equation}
and
\begin{equation}
    \cos(a-b) = \cos(a)\cos(b) + \sin(a)\sin(b).
\end{equation}
By the definition of SLERP in \Eqn{eq:slerp},
\begin{equation}
    \mathbf{z}_\alpha
    =
    \frac{\sin((1-\alpha)\omega)}{\sin\omega}\,\mathbf{z}_0
    +
    \frac{\sin(\alpha\omega)}{\sin\omega}\,\hat{\mathbf{e}}_j.
\end{equation}
Therefore,
\begin{equation}
\begin{aligned}
    \langle \mathbf{z}_\alpha, \hat{\mathbf{e}}_j \rangle
    &= \frac{\sin((1-\alpha)\omega)}{\sin\omega}\,\langle \mathbf{z}_0, \hat{\mathbf{e}}_j \rangle
    + \frac{\sin(\alpha\omega)}{\sin\omega} \\
    &= \frac{1}{\sin\omega}\Big[\sin((1-\alpha)\omega)\cos\omega + \sin(\alpha\omega)\Big] \\
    &= \frac{1}{\sin\omega}\Big[\big(\sin\omega\cos(\alpha\omega) - \cos\omega\sin(\alpha\omega)\big)\cos\omega + \sin(\alpha\omega)\Big] \\
    &= \frac{1}{\sin\omega}\Big[\sin\omega\cos\omega\cos(\alpha\omega) - \cos^2\omega\sin(\alpha\omega) + \sin(\alpha\omega)\Big] \\
    &= \frac{1}{\sin\omega}\Big[\sin\omega\cos\omega\cos(\alpha\omega) + (1-\cos^2\omega)\sin(\alpha\omega)\Big] \\
    &= \frac{1}{\sin\omega}\Big[\sin\omega\cos\omega\cos(\alpha\omega) + \sin^2\omega\sin(\alpha\omega)\Big] \\
    &= \cos\omega\cos(\alpha\omega) + \sin\omega\sin(\alpha\omega) \\
    &= \cos(\omega-\alpha\omega)
    = \cos\!\bigl(\omega(1-\alpha)\bigr),
\end{aligned}
\end{equation}
which proves \Eqn{eq:app-slerp-clean-score}. We can now derive \Eqn{eq:alpha-crit}.

\paragraph{Approximation 1.}
Since $\mathbf{z}_0$ and $\hat{\mathbf{e}}_j$ are independent random points on $\mathbb{S}^{d-1}$, \Eqn{eq:app-one-sided-sphere-tail} implies that they are close to orthogonal with high probability in high dimension. We therefore approximate the initial angle by
\begin{equation}
    \omega \approx \frac{\pi}{2}.
\end{equation}
Substituting this into \Eqn{eq:app-slerp-clean-score} yields
\begin{equation}
    \langle \mathbf{z}_\alpha, \hat{\mathbf{e}}_j \rangle
    \approx
    \cos\!\left(\frac{\pi}{2}(1-\alpha)\right)
    =
    \sin\!\left(\frac{\pi}{2}\alpha\right).
\end{equation}
\paragraph{Approximation 2.}
Now define
\begin{equation}
    M_\alpha := \max_{v \neq j} \langle \mathbf{z}_\alpha, \hat{\mathbf{e}}_v \rangle,
\end{equation}
that is, the maximum similarity among codewords that the simplified generation dynamics does not interpolate toward. Using a union bound together with \Eqn{eq:app-one-sided-sphere-tail}, we obtain
\begin{equation}
    \mathbb{P}\!\left(M_\alpha \geq t \mid \mathbf{z}_\alpha\right)
    \leq
    \sum_{v \neq j}
    \mathbb{P}\!\left(\langle \mathbf{z}_\alpha, \hat{\mathbf{e}}_v \rangle > t \mid \mathbf{z}_\alpha\right)
    \leq
    2(|\mathcal{V}|-1)\exp\!\left(-\frac{d t^2}{2}\right).
\end{equation}

We now solve for the threshold $t_{|\mathcal{V}|,d,\delta}$ such that, with probability at least $1-\delta$, we have $M_\alpha < t_{|\mathcal{V}|,d,\delta}$. Concretely, we solve
\begin{equation}
    \delta
    =
    2(|\mathcal{V}|-1)\exp\!\left(-\frac{d\,t_{|\mathcal{V}|,d,\delta}^2}{2}\right),
\end{equation}
which gives
\begin{equation}
    t_{|\mathcal{V}|,d,\delta}
    =
    \sqrt{
        \frac{2\log\!\bigl(2(|\mathcal{V}|-1)/\delta\bigr)}{d}
    }.
\end{equation}

We can now conclude by defining $\alpha^\star(\delta)$ as the interpolation level after which $\mathbf{z}_\alpha$ is most similar to $\hat{\mathbf{e}}_j$ with probability at least $1-\delta$. A sufficient condition is
\begin{equation}
\begin{aligned}
    \cos\!\bigl((1-\alpha^\star)\omega\bigr)
    &\approx
    \sin\!\left(\frac{\pi}{2}\alpha^\star\right)
    \geq
    t_{|\mathcal{V}|,d,\delta} \\
    &\Longleftrightarrow
    \sin\!\left(\frac{\pi}{2}\alpha^\star\right)
    \geq
    \sqrt{
        \frac{2\log\!\bigl(2(|\mathcal{V}|-1)/\delta\bigr)}{d}
    } \\
    &\Longleftrightarrow
    \alpha^\star(\delta)
    \approx
    \frac{2}{\pi}
    \arcsin\!\left(
        \sqrt{
            \frac{2\log\!\bigl(2(|\mathcal{V}|-1)/\delta\bigr)}{d}
        }
    \right).
\end{aligned}
\end{equation}
This concludes the argument.
\paragraph{Numerical Evaluation of the Critical Threshold.}
\tab{tab:alpha-crit-grid} shows the critical threshold for various vocabulary size $|\mathcal{V}|$, embedding dimension $d$, and confidence parameter $\delta$. As expected, the threshold \emph{decreases} as the embedding dimension increases, which means that the minimum noise level used during training should increase to counteract the increasing sparsity of $\mathbb{S}^{d-1}$. Furthermore, reducing $\delta$ corresponds to increasing the confidence level for similarity to the clean token being the largest, hence it means the minimal noise level is \emph{smaller} than with a larger value of $\delta$.
\begin{table*}[t]
  \caption{$\alpha^\star(\delta)$ for different $|\mathcal{V}|$, $d$, and $\delta$.}
  \label{tab:alpha-crit-grid}
  \centering
  \footnotesize
  \begin{minipage}[t]{0.215\textwidth}
    \centering
    \textit{$|\mathcal{V}|=12$}\\
    \begin{tabular}{@{}lcc@{}}
    \toprule
    $d$ & $\delta=0.1$ & $\delta=0.01$ \\
    \midrule
    256 & 0.132 & 0.158 \\
    512 & 0.093 & 0.111 \\
    768 & 0.076 & 0.090 \\
    1024 & 0.065 & 0.078 \\
    4096 & 0.033 & 0.039 \\
    \bottomrule
    \end{tabular}
  \end{minipage}
  \hfill
  \begin{minipage}[t]{0.215\textwidth}
    \centering
    \textit{$|\mathcal{V}|=50,000$}\\
    \begin{tabular}{@{}lcc@{}}
    \toprule
    $d$ & $\delta=0.1$ & $\delta=0.01$ \\
    \midrule
    256 & 0.213 & 0.231 \\
    512 & 0.149 & 0.161 \\
    768 & 0.121 & 0.131 \\
    1024 & 0.105 & 0.114 \\
    4096 & 0.052 & 0.057 \\
    \bottomrule
    \end{tabular}
  \end{minipage}
  \hfill
  \begin{minipage}[t]{0.215\textwidth}
    \centering
    \textit{$|\mathcal{V}|=100,000$}\\
    \begin{tabular}{@{}lcc@{}}
    \toprule
    $d$ & $\delta=0.1$ & $\delta=0.01$ \\
    \midrule
    256 & 0.219 & 0.236 \\
    512 & 0.153 & 0.165 \\
    768 & 0.125 & 0.134 \\
    1024 & 0.108 & 0.116 \\
    4096 & 0.054 & 0.058 \\
    \bottomrule
    \end{tabular}
  \end{minipage}
\end{table*}
\subsection{Sudoku Setup}
\label{sec:appendix:sudoku-setup}
We format the input as a 1D sequence, starting with the partial grid where the missing digits are encoded with \texttt{0}. We append a \bostoken separator and the solution, and separate rows with \septoken. This representation allows us to train both AR and diffusion models with the same format with a context length of 180. Our vocabulary contains 12 tokens, except for MDLM, which additionally uses a \mask token. When training the diffusion models, the partial grid is never corrupted, and we never penalize the predictions on those positions.

We train all models with Adam for 20k steps, batch size 256, and a learning rate of $3 \times 10^{-4}$ and an \emph{Exponential Moving Average} (EMA) rate of 0.9999. For \method, we consider the linear and cos$^2$ schedules and truncate following \Eqn{eq:alpha-crit} (value in \tab{tab:alpha-crit-grid}).
\subsection{Additional Ablations on Sudoku}
\label{sec:appendix:sudoku-ablations}
\tab{tab:sfm_renorm} ablates two design choices on Sudoku, the noise schedule and the embedding renormalization step. We define the noise schedules in \tab{tab:noise-schedules}.
\begin{table}[t]
    \centering
    \small
    \caption{Analytical noise schedules used on Sudoku, in the convention $\alpha_0 = 0$, $\alpha_1 = 1$.}
    \label{tab:noise-schedules}
    \begin{tabular}{l c}
    \toprule
    Schedule & $\alpha_t$ \\
    \midrule
    Linear & $t$ \\
    Cosine$^2$ & $\sin^2\!\left(\dfrac{\pi t}{2}\right)$ \\
    \bottomrule
    \end{tabular}
\end{table}
\paragraph{Effect of the Noise Schedule.}
Truncating to $\alpha^\star(0.1)$ improves accuracy on both Sudoku (\tab{tab:sudoku}) and TinyGSM (\tab{tab:tinygsm_main}). On Sudoku, the gains are largest for the linear schedule (\tab{tab:sfm_renorm}). Cosine$^2$ also benefits from truncation, but by a smaller margin. Without truncation, Cosine$^2$ outperforms the linear schedule.
\paragraph{Effect of Embedding Renormalization.}
Training without renormalizing the embeddings after each step improves accuracy across nearly all schedules and difficulties (\tab{tab:sfm_renorm}). The gap is largest at the hard difficulty, reaching 7.3 points on Cosine$^2$ without truncation.

\begin{table}[t]
    \caption{Effect of embedding renormalization on \method. Each row compares the same config without vs.\ with embedding re-normalization after every optimizer update. We \textbf{bold} the best result per row and difficulty. Training without re-normalization after each step leads to stronger performance.}
    \label{tab:sfm_renorm}
    \centering
    \small
    \setlength{\tabcolsep}{4pt}
    \begin{tabular}{l ccc | ccc}
    \toprule
    & \multicolumn{3}{c|}{Without renorm} & \multicolumn{3}{c}{With renorm} \\
    \cmidrule(lr){2-4} \cmidrule(lr){5-7}
    Schedule & Easy & Med. & Hard & Easy & Med. & Hard \\
    \midrule
    Linear, no trunc & \textbf{81.5} & \textbf{50.6} & \textbf{14.0} & 75.4 & 46.0 & 11.3 \\
    Linear, trunc $\alpha^\star(0.1)$ & \textbf{94.0} & \textbf{77.6} & \textbf{43.2} & 92.5 & 76.7 & 40.1 \\
    Cosine$^2$, no trunc & 89.9 & \textbf{68.0} & \textbf{36.1} & \textbf{92.1} & 66.3 & 28.8 \\
    Cosine$^2$, trunc $\alpha^\star(0.1)$ & \textbf{93.8} & \textbf{80.7} & \textbf{38.6} & 92.9 & 76.4 & 34.1 \\
    \bottomrule
    \end{tabular}
\end{table}
As discussed in \supp{sec:forced-normalization}, the gradient with respect to a normalized embedding is tangent to the sphere, so each gradient step increases the norm of the embeddings. This is equivalent to a learning-rate decay on the embedding table. Empirically, such annealing improves performance on Sudoku (\tab{tab:sfm_renorm}).
\subsection{Bootstrap Confidence Intervals on TinyGSM}
\label{sec:appendix:bootstrap}
The error bars on the TinyGSM accuracy plots are $95\%$ percentile bootstrap confidence intervals \citep{efron1979bootstrap, raschka2022confidence}. The GSM8K test split contains $N = 1{,}319$ problems, each with a generated solution that is either correct or incorrect. We resample the $N$ outcomes with replacement $B = 1{,}000$ times and recompute the accuracy on every resample. The point estimate is the mean of the $B$ accuracies. The error bar runs from the $2.5\%$ quantile of the $B$ resample accuracies to the $97.5\%$ quantile, namely the $2.5$th and $97.5$th percentiles of the empirical bootstrap accuracy distribution.

\subsection{Additional Ablations on TinyGSM}
\label{sec:appendix:tinygsm-ablations}
\tab{tab:sfm_tinygsm_all_schedules} sweeps the truncation hyperparameter $\alpha_{\text{trunc}}$ on the linear schedule, with the standard DiT backbone trained for 250k steps. Truncating the noise schedule improves the accuracy from $1.21\%$ at $\alpha_{\text{trunc}} = 0$ to $7.73\%$ at $\alpha_{\text{trunc}} = 0.9$. The thresholds from \Eqn{eq:alpha-crit} for $d = 768$ and $|\mathcal{V}| \approx 50\text{k}$, $\alpha^\star(0.1) = 0.879$ and $\alpha^\star(0.01) = 0.869$, perform similarly to the best run ($\alpha_{\text{trunc}} = 0.9$).

\begin{table}[t]
    \caption{Effect of the truncation hyperparameter $\alpha_{\text{trunc}}$ on \method accuracy on GSM8K (trained on TinyGSM). Linear noise schedule, standard DiT backbone, 250k steps. The values $0.879$ and $0.869$ are the thresholds $\alpha^\star(\delta)$ from \Eqn{eq:alpha-crit} at $\delta = 0.1$ and $\delta = 0.01$ respectively. \textbf{Bold} marks the best result.}
    \label{tab:sfm_tinygsm_all_schedules}
    \centering
    \small
    \begin{tabular}{cc}
    \toprule
    $\alpha_{\text{trunc}}$ & Accuracy (\%) \\
    \midrule
    $0.0$    & $1.21$ \\
    $0.8$    & $4.62$ \\
    $0.869$  & $7.66$ \\
    $0.879$  & $7.43$ \\
    $0.9$    & $\mathbf{7.73}$ \\
    \bottomrule
    \end{tabular}
\end{table}

\subsection{Additional Sampling Results on TinyGSM}
\label{sec:appendix:tinygsm-additional}
\begin{figure}[H]
    \centering
    \includegraphics[width=0.65\linewidth]{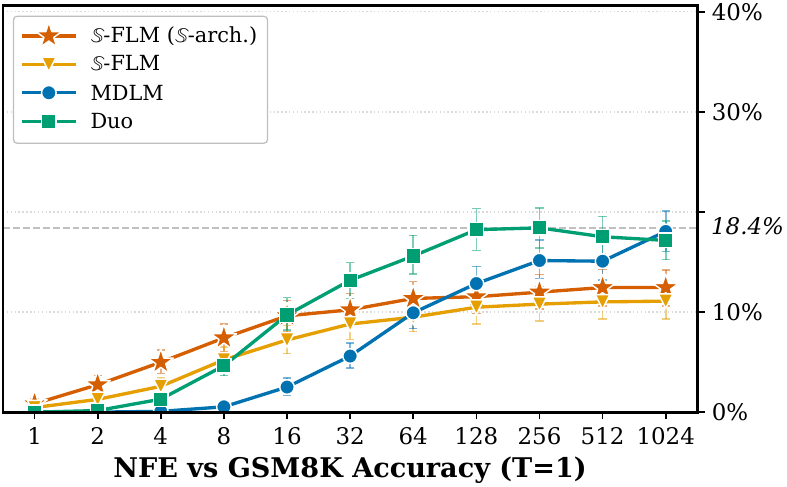}
    \caption{GSM8K accuracy vs.\ NFE at $T = 1$ for \method (\arch and standard DiT), MDLM, and Duo. Same data as \Fig{fig:tinygsm-T1p0} (right).}
    \label{fig:appendix-tinygsm-arch-baselines-T1p0}
\end{figure}

\begin{figure}[H]
    \centering
    \includegraphics[width=0.65\linewidth]{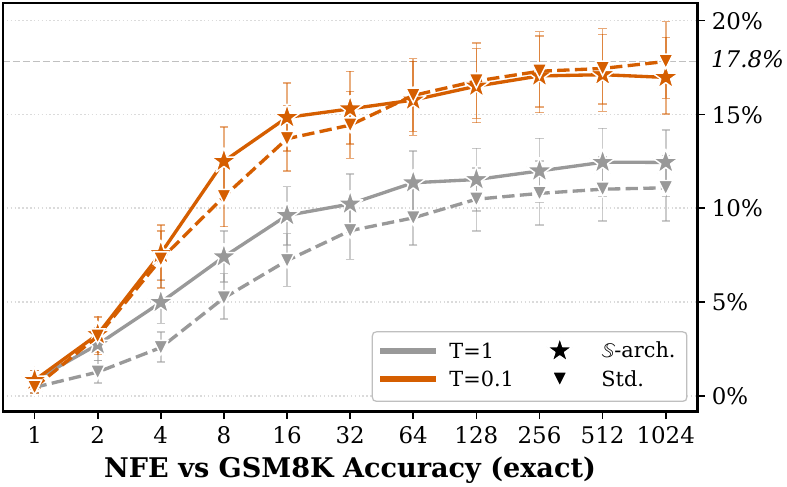}
    \caption{GSM8K accuracy vs.\ NFE under exact decoding for \method with \arch (Norm.) and standard DiT (Std.), at $T = 1$ (grey) and $T = 0.1$ (orange). Lowering the temperature improves accuracy by about six points for both backbones.}
    \label{fig:appendix-tinygsm-arch-x-temp}
\end{figure}

\begin{figure}[H]
    \centering
    \includegraphics[width=0.65\linewidth]{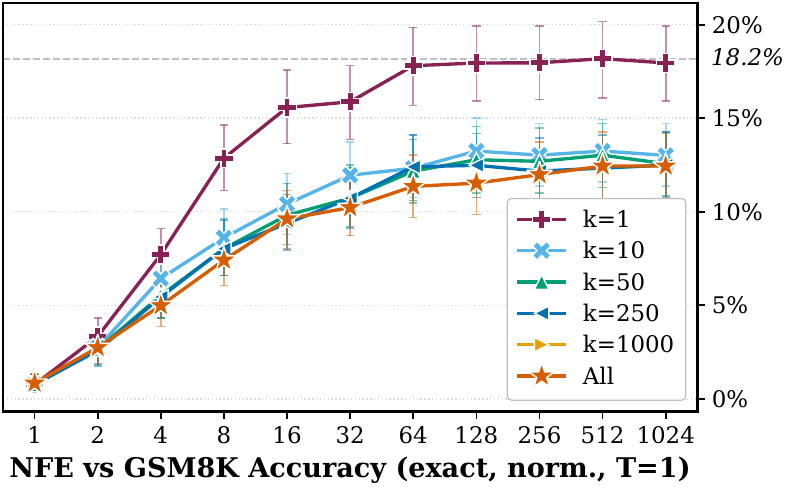}
    \caption{GSM8K accuracy vs.\ NFE for \method (\arch) at $T = 1$, sweeping the top-$k$ truncation of the predicted velocity field at each Euler step. Top-$1$ reaches $18.0\%$, while $k \geq 10$ all plateau near $12\%$, matching unrestricted decoding.}
    \label{fig:appendix-tinygsm-velocity-topk-norm-T1p0}
\end{figure}

For models trained on TinyGSM, we sweep architecture, sampling temperature, and velocity decoding, and report GSM8K accuracy. \Fig{fig:appendix-tinygsm-arch-baselines-T1p0} replots \Fig{fig:tinygsm-T1p0} (right). At $T = 0.1$, Duo reaches $36.0\%$ and MDLM about $33\%$, while \method plateaus near $18\%$. \Fig{fig:appendix-tinygsm-arch-x-temp} compares the \arch and standard DiT backbones at both temperatures, with $T = 0.1$ improving GSM8K accuracy by about six points for both. \Fig{fig:appendix-tinygsm-velocity-topk-norm-T1p0} sweeps the top-$k$ truncation of velocity decoding for \arch at $T = 1$. Top-$1$ reaches $18.0\%$, while $k \geq 10$ plateau near $12\%$, matching unrestricted decoding.

\newpage

\begin{figure}[H]
    \centering
    \begin{subfigure}[t]{0.49\textwidth}
        \centering
        \includegraphics[width=\linewidth]{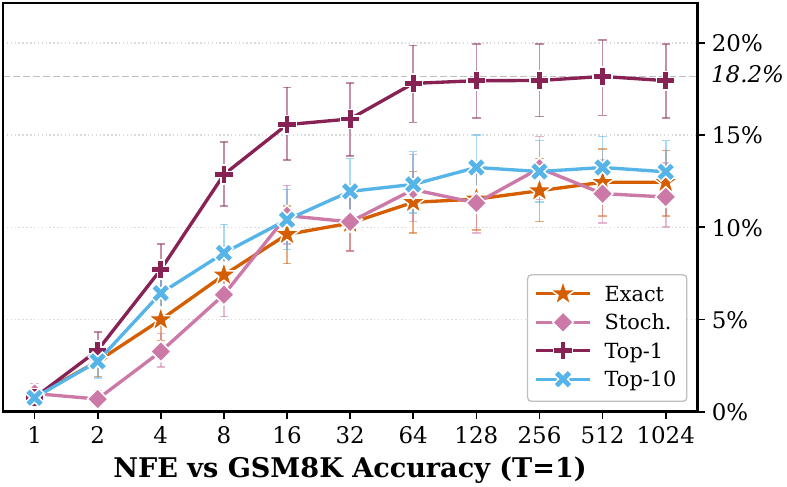}
    \end{subfigure}\hfill
    \begin{subfigure}[t]{0.49\textwidth}
        \centering
        \includegraphics[width=\linewidth]{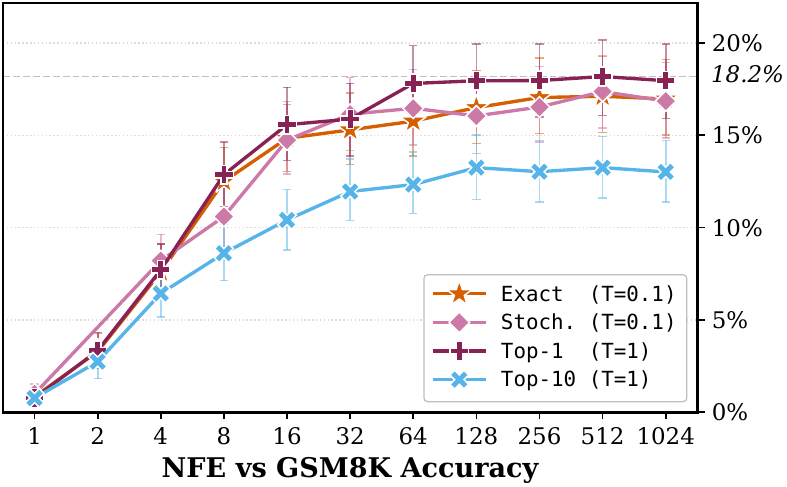}
    \end{subfigure}
    \caption{GSM8K accuracy vs.\ NFE for \method (\arch) under exact velocity, stochastic, top-$1$, and top-$10$ decoding. \textbf{(left)} Sampling temperature $T = 1$. Exact velocity and stochastic decoding plateau near $12\%$, while top-$1$ reaches $18.0\%$. \textbf{(right)} Sampling temperature $T = 0.1$. All four schemes plateau within one point of $18.0\%$. Top-$1$ decoding ($T = 1$) outperforms low-temperature stochastic decoding.}
    \label{fig:appendix-tinygsm-velocity-topk-T0p1-vs-T1p0}
\end{figure}

\subsection{Gen.\ PPL / Entropy Frontier Protocol}
\label{sec:appendix:owt-protocol}
We adopt the same setup as in PGM \citep{deschenaux2025partitiongenerativemodelingmasked}. For each method (MDLM, Duo, FLM, and \method with exact, stochastic, and top-$1$ velocity decoding), we sweep the sampling temperature $T \in \{0.50, 0.55, \ldots, 1.20\}$ ($15$ evenly-spaced values) and the number of function evaluations NFE $\in \{32, 64, 128, 256, 512, 1024\}$. For every $(T, \mathrm{NFE})$ cell, we draw $N = 512$ samples of length $L = 1024$ from the OpenWebText-trained model and compute two scalars per sample.
\paragraph{Generative Perplexity.} We score each generated sample with a pretrained GPT-2-large reference model \citep{deschenaux2025partitiongenerativemodelingmasked} and report the per-sample average
\begin{equation}
\mathrm{PPL}_{\mathrm{gen}} = \exp\!\left(-\frac{1}{L} \sum_{i = 1}^{L} \log p_{\mathrm{GPT2}}\!\left(x_i \mid x_{<i}\right)\right),
\end{equation}
masking tokens at or after the first end-of-text. We average across the $N$ samples in the cell.
\paragraph{Unigram Entropy.} To flag degenerate (repetitive) outputs, we compute the per-cell mean unigram entropy \citep{deschenaux2025partitiongenerativemodelingmasked}
\begin{equation}
H = -\frac{1}{N} \sum_{n = 1}^{N} \sum_{v \in \mathcal{V}} \frac{c(v, x^{(n)})}{L} \log \frac{c(v, x^{(n)})}{L},
\end{equation}
where $c(v, x^{(n)})$ counts the occurrences of token $v$ in sample $x^{(n)}$.
\paragraph{Visible Window.} Each cell contributes one $(H, \mathrm{PPL}_{\mathrm{gen}})$ point. We restrict the visible window to $4.5 \le H \le 6.0$ nats and $\mathrm{PPL}_{\mathrm{gen}} \le 500$ to drop collapsed and degenerate configurations. \method with top-$1$ velocity does not depend on the temperature, so it appears as a single marker rather than a curve.

\subsection{Additional Results on OpenWebText}
\label{sec:appendix:owt-additional}
\begin{figure}[H]
    \centering
    \begin{subfigure}[t]{0.48\textwidth}
        \centering
        \includegraphics[width=\linewidth]{figures/owt/owt_gen_ppl_vs_entropy_nfe32.pdf}
    \end{subfigure}\hfill
    \begin{subfigure}[t]{0.48\textwidth}
        \centering
        \includegraphics[width=\linewidth]{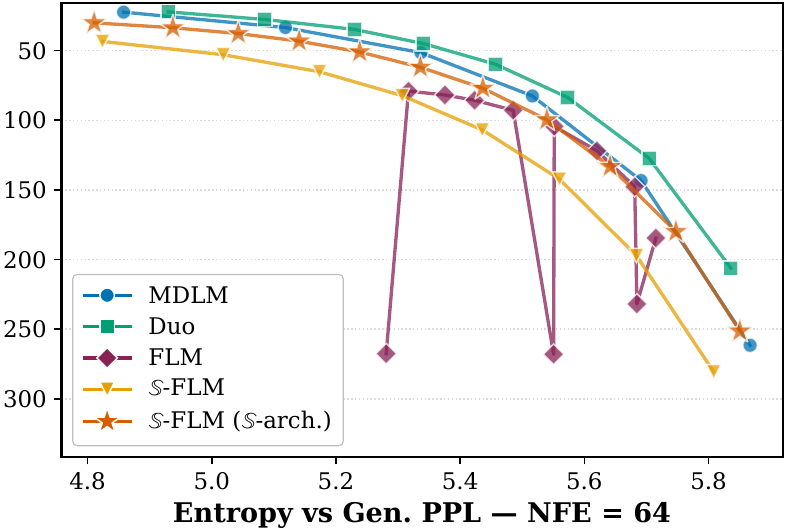}
    \end{subfigure}

    \vspace{0.6em}

    \begin{subfigure}[t]{0.48\textwidth}
        \centering
        \includegraphics[width=\linewidth]{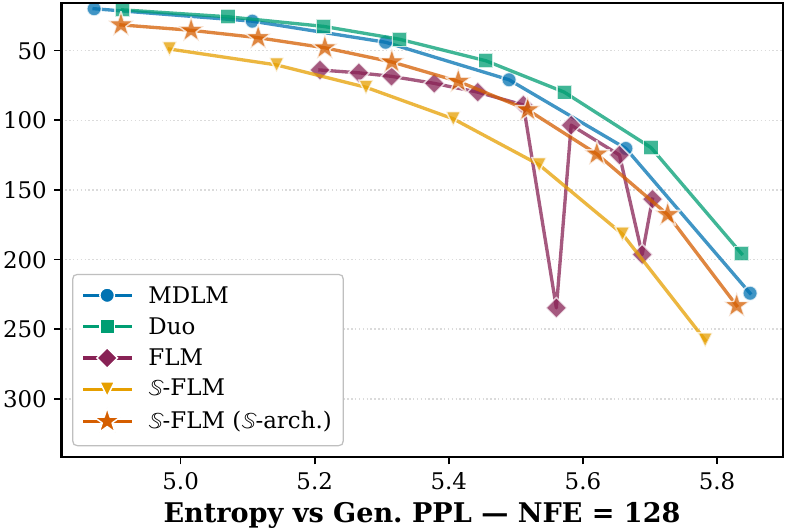}
    \end{subfigure}\hfill
    \begin{subfigure}[t]{0.48\textwidth}
        \centering
        \includegraphics[width=\linewidth]{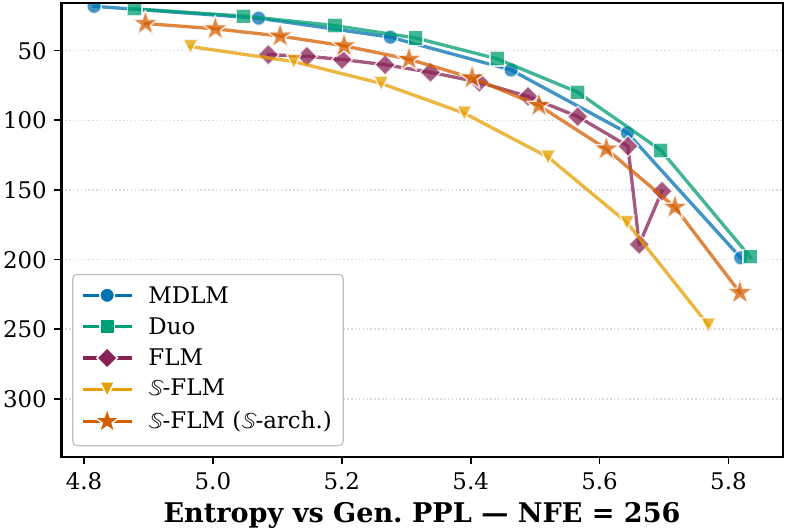}
    \end{subfigure}

    \vspace{0.6em}

    \begin{subfigure}[t]{0.48\textwidth}
        \centering
        \includegraphics[width=\linewidth]{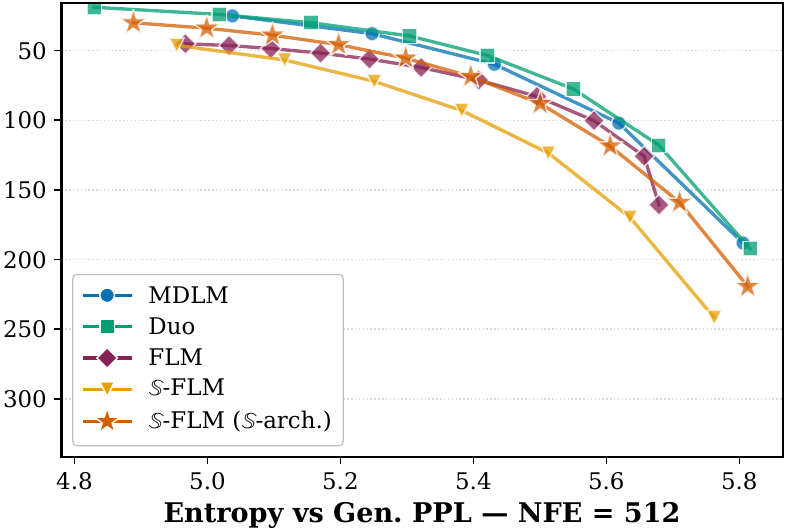}
    \end{subfigure}\hfill
    \begin{subfigure}[t]{0.48\textwidth}
        \centering
        \includegraphics[width=\linewidth]{figures/owt/owt_gen_ppl_vs_entropy_nfe1024.pdf}
    \end{subfigure}
    \caption{OpenWebText Gen.\ PPL versus per-sample unigram entropy for NFE $\in \{32, 64, 128, 256, 512, 1024\}$. Each curve sweeps a sampling-temperature schedule. \method (exact velocity and stochastic decoding) traces the Pareto frontier across every NFE budget, with the largest margin over MDLM, Duo, and FLM at low NFE.}
    \label{fig:appendix-owt-pareto-all}
\end{figure}

\Fig{fig:appendix-owt-pareto-all} shows the Gen.\ PPL / Entropy front for NFEs $\in \{32, 64, 128, 256, 512, 1024\}$.

\newpage

\section{Extended Related Work}
\label{sec:appendix:related-work}
\method differs from prior work in three main ways. (1) It operates in continuous rather than discrete space, (2) defines the flow on the hypersphere rather than in Euclidean space, and (3) learns embeddings end-to-end instead of relying on pre-trained representations.
\paragraph{Discrete Diffusion Language Models.}
Discrete diffusion models for text \citep{austin2023structureddenoisingdiffusionmodels, campbell2022continuoustimeframeworkdiscrete, lou2024discretediffusionmodelingestimating, sahoo2024simpleeffectivemaskeddiffusion, gat2024discreteflowmatching, campbell2024generativeflowsdiscretestatespaces, sahoo2025diffusionduality, deschenaux2026diffusiondualitychapterii, shi2025simplifiedgeneralizedmaskeddiffusion, nie2025scalingmaskeddiffusionmodels, sahoo2025esotericlanguagemodels, vonrütte2025generalizedinterpolatingdiscretediffusion, deschenaux2025partitiongenerativemodelingmasked, vonruette2026scalingbehaviordiscretediffusion, sahoo2026scalingmaskeddiffusionlanguage, arriola2025blockdiffusioninterpolatingautoregressive} approach AR in Gen.\ PPL while enabling parallel generation, but require updating tokens independently at inference for tractability. Furthermore, because the sampling steps are not differentiable, gradient-based guidance \citep{dhariwal2021diffusion, ho2022classifierfreediffusionguidance, nisonoff2024unlocking, schiff2025simpleguidancemechanismsdiscrete} is harder than in the continuous case.
\paragraph{Continuous Diffusion for Language Modeling.}
Instead of operating in the discrete space of tokens, certain prior work use Gaussian diffusion \citep{sohldickstein2015deepunsupervisedlearningusing, ho2020denoisingdiffusionprobabilisticmodels} for language modeling. One approach is to apply Gaussian diffusion to embeddings and train end-to-end with CE \citep{li2022diffusionlmimprovescontrollabletext, dieleman2022continuousdiffusioncategoricaldata, gulrajani2023likelihoodbaseddiffusionlanguagemodels}. Alternatively, others regress onto pre-trained embeddings \citep{strudel2022selfconditionedembeddingdiffusion, lovelace2022latentdiffusionlanguagegeneration, shen2026codarcontinuousdiffusionlanguage}. The latter requires two training stages, and the sample quality is capped by the pre-trained embeddings. Instead of embeddings, recent work add Gaussian noise to one-hot or simplex representations of the tokens \citep{chen2022analogbits, han2023ssdlmsemiautoregressivesimplexbaseddiffusion, mahabadi2023tess, stark2024dirichletflowmatching, lee2026onesteplanguagemodelingcontinuous, roos2026categoricalflowmaps, potaptchik2026discreteflowmaps}. This approach requires storing dense arrays $L \times |\mathcal{V}|$ during training and sampling. Separately, Loopholing discrete diffusion \citep{jo2025loopholingdiscretediffusion} allows discrete diffusion models to propagate the latent representation of the denoiser across sampling steps. \method is defined on the latent hypersphere. Therefore, it does not require the materialization of dense arrays $L \times |\mathcal{V}|$ during training. We inject noise by rotating embeddings rather than adding Gaussian noise.
\paragraph{Representation Learning on the Hypersphere.}
Hyperspherical representations are common in contrastive learning, where uniform spread in $\mathbb{S}^{d-1}$ correlates with strong downstream performance \citep{wang2020understandingcontrastiverepresentationlearning}. Comparing word embeddings with cosine similarity performs better than using the Euclidean distance in high dimension \citep{mikolov2013efficientestimationwordrepresentations, pennington2014glove}. Thus, cosine similarity underpins a large part of neural retrieval systems \citep{reimers2019sentencebertsentenceembeddings, karpukhin2020densepassageretrieval}. Training Variational Autoencoders with a latent prior in $\mathbb{S}^{d-1}$ can be more stable than with Gaussian priors \citep{davidson2018hypersphericalvariationalautoencoders, xu2018sphericallatentspacesvae}. Normalizing activations and weights to $\mathbb{S}^{d-1}$ can also improve the stability of AR models \citep{loshchilov2024ngptnormalizedtransformerrepresentation}. In Riemannian manifolds, predicting the clean endpoint can outperform regressing the velocity field \citep{eijkelboom2025variationalflowmatchinggraph, zaghen2026riemannianvariationalflowmatching}. Previous work extended CNFs and score-based diffusion to Riemannian Manifolds \citep{lou2020neuralmanifoldode, mathieu2020riemanniancnf, debortoli2022riemannianscorebased, lou2023scalingriemanniandiffusion}. However, they typically assume data already reside on the manifold. In contrast, we learn the token representation and the velocity on $\mathbb{S}^{d-1}$ jointly. Fisher-Flow \citep{davis2024fisherflowmatching} maps one-hot vector on the positive orthant of the hypersphere, but this representation does not perform well on language modeling with large vocabularies \citep{jo2025continuousdiffusionmodellanguage}.

\newpage

\end{document}